\documentclass[runningheads]{llncs}

\usepackage{graphicx}
\usepackage{comment}
\usepackage{amsmath,amssymb} 
\usepackage{color}

\usepackage{wrapfig}
\usepackage{subfigure}
\usepackage{footnote}
\usepackage{hyperref}

\usepackage[font=small,labelfont=bf]{caption}

\graphicspath{{images/}}
\usepackage[export]{adjustbox}
\usepackage{array}

\usepackage[width=122mm,left=12mm,paperwidth=146mm,height=193mm,top=12mm,paperheight=217mm]{geometry}

\begin{document}

\pagestyle{headings}
\mainmatter
\def\ECCVSubNumber{3568}  

\def\ie{\textit{i.e.}}

\title{Do Not Mask What You Do Not Need to Mask: a Parser-Free Virtual Try-On} 

\titlerunning{A Parser-Free Virtual Try-On}
%

\author{Thibaut Issenhuth \inst{1} \and
J\'er\'emie Mary \inst{1} \and
Cl\'ement Calauz\`enes\inst{1}}
\authorrunning{T. Issenhuth et al.}
%
\institute{Criteo AI Lab, Paris, France \\
\email{\{t.issenhuth,j.mary,c.calauzenes\}@criteo.com}}

\maketitle

\begin{abstract}
  The 2D virtual try-on task has recently attracted a great interest from the research community, for its direct potential applications in online shopping as well as for its inherent and non-addressed scientific challenges. This task requires fitting an in-shop cloth image on the image of a person, which is highly challenging because it involves cloth warping, image compositing, and synthesizing.
  Casting virtual try-on into a supervised task faces a difficulty: available datasets are composed of pairs of pictures (cloth, person wearing the cloth). Thus, we have no access to ground-truth when the cloth on the person changes.
  State-of-the-art models solve this by masking the cloth information on the person with both a human parser and a pose estimator. Then, image synthesis modules are trained to reconstruct the person image from the masked person image and the cloth image.
 This procedure has several caveats: firstly, human parsers are prone to errors; secondly, it is a costly pre-processing step, which also has to be applied at inference time; finally, it makes the task harder than it is since the mask covers information that should be kept such as hands or accessories.
 In this paper, we propose a novel student-teacher paradigm where the teacher is trained in the standard way (reconstruction) before guiding the student to focus on the initial task (changing the cloth). The student additionally learns from an adversarial loss, which pushes it to follow the distribution of the real images. Consequently, the student exploits information that is masked to the teacher. A student trained without the adversarial loss would not use this information. Also, getting rid of both human parser and pose estimator at inference time allows obtaining a real-time virtual try-on.
\end{abstract}

\section{Introduction}

\setlength{\belowcaptionskip}{-10pt}
\addtolength{\tabcolsep}{-6pt}
\begin{figure}[ht]
\fontsize{8}{8}\selectfont
\centering
\begin{tabular*}{340pt}{@{\extracolsep{\fill}}cccccc}
Reference & Target & Human & CP-VTON & T-WUTON & S-WUTON \\
person & cloth & parsing &  & (ours) & (ours) \\ [-0.2ex]
\subfigure{\includegraphics[width=0.16\linewidth]{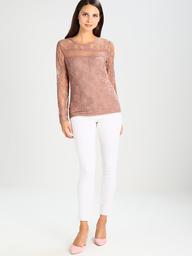}} &
\subfigure{\includegraphics[width=0.16\linewidth]{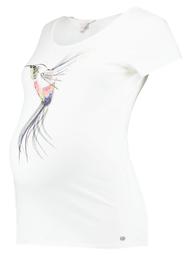}} &
\subfigure{\includegraphics[width=0.16\linewidth]{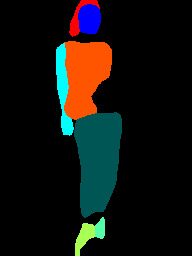}} &
\subfigure{\includegraphics[width=0.16\linewidth]{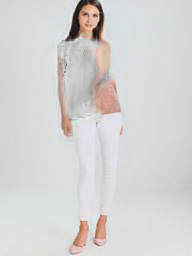}} &
\subfigure{\includegraphics[width=0.16\linewidth]{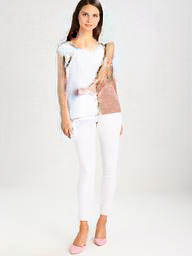}} &
\subfigure{\includegraphics[width=0.16\linewidth]{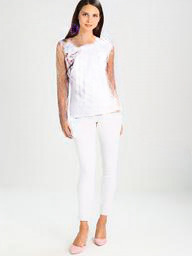}}\\ [-2.6ex]
\subfigure{\includegraphics[width=0.16\linewidth]{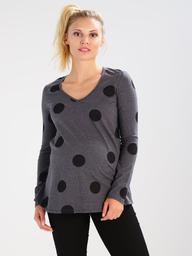}} &
\subfigure{\includegraphics[width=0.16\linewidth]{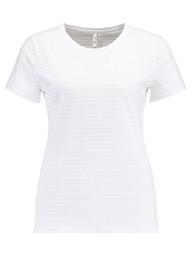}} &
\subfigure{\includegraphics[width=0.16\linewidth]{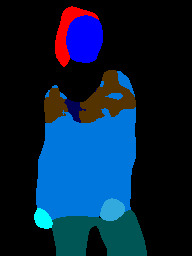}} &
\subfigure{\includegraphics[width=0.16\linewidth]{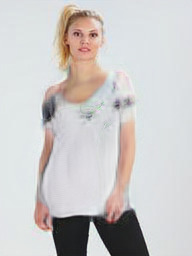}} &
\subfigure{\includegraphics[width=0.16\linewidth]{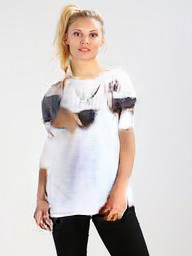}} &
\subfigure{\includegraphics[width=0.16\linewidth]{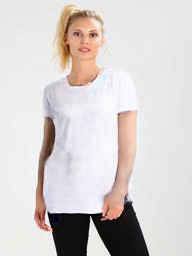}}\\ [-2.6ex]
\subfigure{\includegraphics[width=0.16\linewidth]{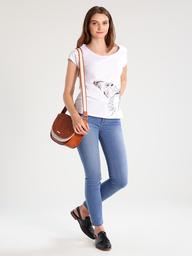}} &
\subfigure{\includegraphics[width=0.16\linewidth]{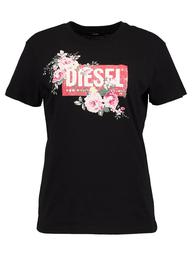}} &
\subfigure{\includegraphics[width=0.16\linewidth]{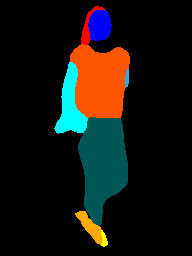}} &
\subfigure{\includegraphics[width=0.16\linewidth]{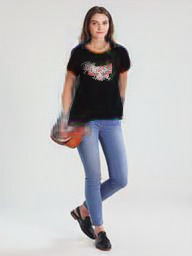}} &
\subfigure{\includegraphics[width=0.16\linewidth]{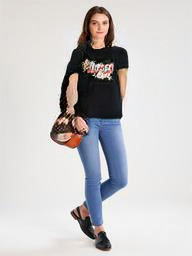}} &
\subfigure{\includegraphics[width=0.16\linewidth]{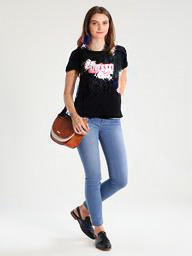}}\\ [-0.6ex]
\end{tabular*}
\caption{\label{fig:introduction} Typical failure cases of the human parser. On the two first rows, it does not segment the person properly. On the third row, it masks the handbag which we would like to preserve in a virtual try-on. CP-VTON and our T-WUTON, which rely on the parsing information, are not robust to a bad parsing. However, the student model S-WUTON which is distilled from the human parser, pose estimator and T-WUTON, can preserve the person's attributes and does not rely on the parsing information.}
\end{figure}
\addtolength{\tabcolsep}{6pt}
A photo-realistic virtual try-on system would provide a significant improvement for online shopping. Whether used to create catalogs of new products or to propose an immersive environment for shoppers, it could impact e-commerce and open the door for automated image-editing possibilities.

Earlier work  addresses this challenge using  3D measurements and model-based methods \cite{guan2012drape,hahn2014subspace,pons2017clothcap}. However, these are, by nature, computationally intensive and require expensive material, which would not be acceptable at scale for shops. Recent works aim to leverage deep generative models to tackle the virtual try-on problem \cite{dong2019towards,han2018viton,jetchev2017conditional,wang2018toward}. CAGAN \cite{jetchev2017conditional} is a U-net based Cycle-GAN \cite{isola2017image} approach. However, this method fails to generate realistic results since such networks cannot handle large spatial deformations. In VITON \cite{han2018viton}, the authors recast the virtual try-on as a supervised task. They propose to use a human parser and a pose estimator to mask the cloth in the person image and construct an agnostic person representation $p^{\star}$. The human parser allows segmenting the upper-body and the cloth, while the pose estimator locates the keypoints (\textit{i.e.} shoulders, wrists, etc.) of the person.  Then, with $p^{\star}$ and the image of the original cloth $c$ on a white background, they train a model in a fully supervised fashion to reconstruct $p$. Namely, they propose a coarse-to-fine synthesis strategy with shape context matching algorithm \cite{belongie2002shape} to warp the cloth on the target person. To improve this model, CP-VTON \cite{wang2018toward} incorporates a convolutional geometric matcher \cite{rocco2017convolutional}, which learns geometric deformations (\textit{i.e.} thin-plate spline transform  \cite{bookstein1989principal}) that align the cloth with the person. State-of-the-art models are based on the supervised formulation of the virtual try-on task, which has some drawbacks. Human parsers and pose estimators are trained on other datasets and thus fail in some situations (see Fig. \ref{fig:introduction}, two first rows). Retraining them on fashion datasets would require similar labels of semantic segmentation or unsupervised domain adaptation methods. Even though they would still be imperfect. Moreover, for a virtual try-on, one wants to preserve person's attributes like handbags or jewels. When constructing $p^{\star}$, these person's attributes are masked and can not be preserved, such as the partially masked handbag on the third row of Fig. \ref{fig:introduction}. Finally, the human parsing and pose estimation are the wall clock bottleneck of the
pipeline.

In our work, we distill \cite{hinton2015distilling} the standard pipeline of virtual try-on composed of human parser, pose estimator, and synthesis modules in the synthesis modules. Namely, we train a student synthesizer with the outputs of a pre-trained standard virtual try-on pipeline. To force  the student to use information that is masked to the teacher, we also train the student with an adversarial loss.  The distillation process allows us to remove the need for human parsing and pose estimation at inference time, which improves image quality and speeds up the computations from 6FPS to 77FPS. In Fig. \ref{fig:introduction}, we show visual results of a baseline CP-VTON, our teacher model T-WUTON and our student model S-WUTON. Since S-WUTON does not rely on human parsing, it is robust to parsing errors and preserves a person's attributes such as fingers or handbags.

Additionally, to build an efficient teacher model, we propose an improved architecture for virtual try-on, a Warping U-Net for a Virtual Try-On (WUTON). Our architecture is composed of two modules: a convolutional geometric matcher \cite{rocco2017convolutional} and a U-net generator with a siamese encoder, where the former warps the feature maps of the latter. The architecture is trained end-to-end, which leads to high-quality synthesized images.

We demonstrate the benefit of our method with several experiments on a virtual try-on dataset, with quantitative and visual results, and a user study.

\section{Problem statement and related work}
Given the 2D images $p \in \mathbb{R}^{h\times w \times 3}$ of a person and $c \in \mathbb{R}^{h\times w \times3}$ of a clothing item, we want to generate the image $\Tilde{p} \in \mathbb{R}^{h\times w\times 3}$ where a person $p$ wears the cloth $c$. The task can be separated in two parts : the geometric deformation $T$ required to align $c$ with $p$, and the refinement that fits the aligned cloth $\Tilde{c} = T(c)$ on $p$. These two sub-tasks can be modelled with learnable neural networks, \textit{i.e.} spatial transformers networks $STN$ \cite{jaderberg2015spatial,rocco2017convolutional} that output parameters $\theta = STN(p,c)$ of geometric deformations, and conditional generative networks $G$ that give $\Tilde{p} = G(p,c,\theta)$.

Because it would be costly to construct a dataset with $\{(p,c),\Tilde{p}\}$ triplets, previous works \cite{han2018viton,wang2018toward} propose to use an agnostic person representation $p^\star \in \mathbb{R}^{h\times w\times c}$ where the clothing items in $p$ are hidden but identity and shape of the persons are preserved. $p^\star$ is built with pre-trained human parsers and pose estimators : $p^\star = h(p) $. These triplets $\{(p^\star,c),p\}$ allow to train for reconstrution. $(p^\star,c)$ are the inputs, $\Tilde{p}$ the output and $p$ the ground-truth. We finally have the conditional generative process :
\begin{equation}
\label{eq:pb_statement}
    \Tilde{p} = G(\underbrace{h(p)}_{\text{agnostic person}},\underbrace{c}_{\text{cloth}},\underbrace{STN(h(p),c)}_{\text{geometric transform}})
\end{equation}
Although it eases the training of $G$, $h$ is a bottleneck in the virtual try-on pipeline. We will show that we can train a student model with synthetic triplets $\{(p,c),\Tilde{p}\}$, where $\Tilde{p}$ comes from our pre-trained teacher generative model in Eq.~ \ref{eq:pb_statement}. This allows to remove the need for $h$ at inference time for the student model:
\begin{equation}
    \hat{p} = G_s(\underbrace{p}_{\text{original person}},\underbrace{c}_{\text{cloth}},\underbrace{STN_s(p,c)}_{\text{geometric transform}})
\end{equation}
where $G_s$ and $STN_s$ are the student modules and $\hat{p}$ the generated image.
\paragraph{\textbf{Conditional image generation.}} Generative models for image synthesis have shown impressive results with adversarial training \cite{goodfellow2014generative}.
Combined with deep networks \cite{radford2015unsupervised}, this approach has been extended to conditional image generation in \cite{mirza2014conditional} and performs increasingly well on a wide range of tasks, from image-to-image translation \cite{isola2017image,zhu2017unpaired} to video editing \cite{shetty2018adversarial}. However, as noted in \cite{mejjati2018unsupervised}, these models cannot handle large spatial deformations and fail to modify the shape of objects, which is necessary for a virtual try-on.

\paragraph{\textbf{Appearance transfer.}} Close to the virtual try-on task, some research focus on human appearance transfer. Given two images of different persons, the goal is to transfer the appearance of a part  of the person A on the person B. Approaches using pose and appearance disentanglement \cite{lorenz2019unsupervised,ma2018disentangled} fit this task but others are specifically designed for it. SwapNet \cite{raj2018swapnet} is a dual path network which generates a new human parsing of the reference person and region of interest pooling to transfer the texture. In \cite{wu2018m2e}, the method relies on DensePose information \cite{alp2018densepose}, which provides a 3D surface estimation of a human body, to perform a warping and align the two persons. The transfer is then done with segmentation masks and refinement networks. However, the warping relies on matching source and target pose, which is not feasible for the virtual try-on task.
\paragraph{\textbf{Virtual try-on.}} Most of the approaches for a virtual try-on system come from computer graphics and rely on 3D measurements or representations. Drape \cite{guan2012drape} learns a deformation model to render clothes on 3D bodies of different shapes. In \cite{hahn2014subspace}, Hahn et al. use subspace methods to accelerate physics-based simulations and generate realistic wrinkles.  ClothCap \cite{pons2017clothcap} aligns a 3D cloth-template to each frame of a sequence of 3D scans of a person in motion. However, the use of 3D scans is expensive and thus not doable for online users.

The task we are interested in is the one introduced in CAGAN \cite{jetchev2017conditional} and further studied by VITON \cite{han2018viton} and CP-VTON \cite{wang2018toward}, which we defined in the problem statement. In CAGAN \cite{jetchev2017conditional}, Jetchev et al. propose a cycle-GAN approach that requires three images as input: the reference person, the cloth worn by the person and the target in-shop cloth. Thus, it limits its practical uses. To facilitate the task, VITON \cite{han2018viton} introduces the supervised formulation of the virtual try-on, as described above. Their pipeline separates the task in sub-tasks: constructing the agnostic person representation (\textit{i.e.} mask the area to replace but preserve body shape), warping the cloth and compositing the final image. Based on the agnostic person representation $p^{\star}$ and the cloth image $c$, the VITON model performs a generative composition between the warped cloth and a coarse result. The warping is done with a non-parametric geometric transform \cite{belongie2002shape}. To improve this model, CP-VTON \cite{wang2018toward} incorporates a learnable geometric matcher $STN$ \cite{rocco2017convolutional}. The $STN$ is trained to align $c$ on $p$ with a L1 loss on paired images. However, the L1 loss is overwhelmed with the white background and the solid color parts of clothes. Thus, it faces difficulties to align patterns and to preserve inner structure of the cloth. In VTNFP \cite{Yu_2019_ICCV} and ClothFlow \cite{Han_2019_ICCV}, a module generating the new human parsing is added. It allows to better preserve body parts and edges, but at an increased computational cost. Moreover, ClothFlow \cite{Han_2019_ICCV}  replaces the TPS warping by a dense flow from the target cloth to the person. All these recent works \cite{han2018viton,wang2018toward,Han_2019_ICCV,Yu_2019_ICCV} rely on pre-trained human parser and pose estimator.

Recent work MG-VTON \cite{dong2019towards} extends the task to a multi-pose virtual try-on system, where they also change the pose of the reference person. Similarly to \cite{dong2018soft,Yu_2019_ICCV,dong2019towards}, they add a module generating the new human parsing, based on input and target pose information.

\section{Our approach}
Our task is to build a virtual try-on system that is able to fit a given in-shop cloth on a reference person. In this work, we build a virtual try-on that does not rely on a human parser nor a pose estimator for inference. To do so, we use a teacher-student approach to distill the standard virtual try-on pipeline composed of human parser, pose estimator, and synthesis module in the synthesis module.

In Section \ref{sec:architecture}, we detail the architecture of our synthesis module WUTON. It is trainable end-to-end and composed of two existing modules: a convolutional geometric matcher $STN$ \cite{rocco2017convolutional} and a U-net \cite{ronneberger2015u} with siamese encoder whose skip connections from the cloth encoder to the decoder are deformed by $STN$. We then explain its training procedure in the standard supervised setting, which gives the teacher T-WUTON.

We finally explain our distillation process. Once the first generative model is trained, the pipeline $\{h,\text{T-WUTON}\}$ becomes a teacher model for a student model S-WUTON by constructing synthetic triplets $\{(p,c),\Tilde{p}\}$. These serve to supervise the training of S-WUTON, which hence does not need a human parser to pre-process the image and construct the agnostic person representation. Importantly, S-WUTON also  learns from an adversarial loss so it does not only follow the teacher's distribution and it can learn to preserve a person's attributes.

\subsection{WUTON architecture}
\label{sec:architecture}
\begin{figure*}[ht]
\centering
    \includegraphics[width=\linewidth]{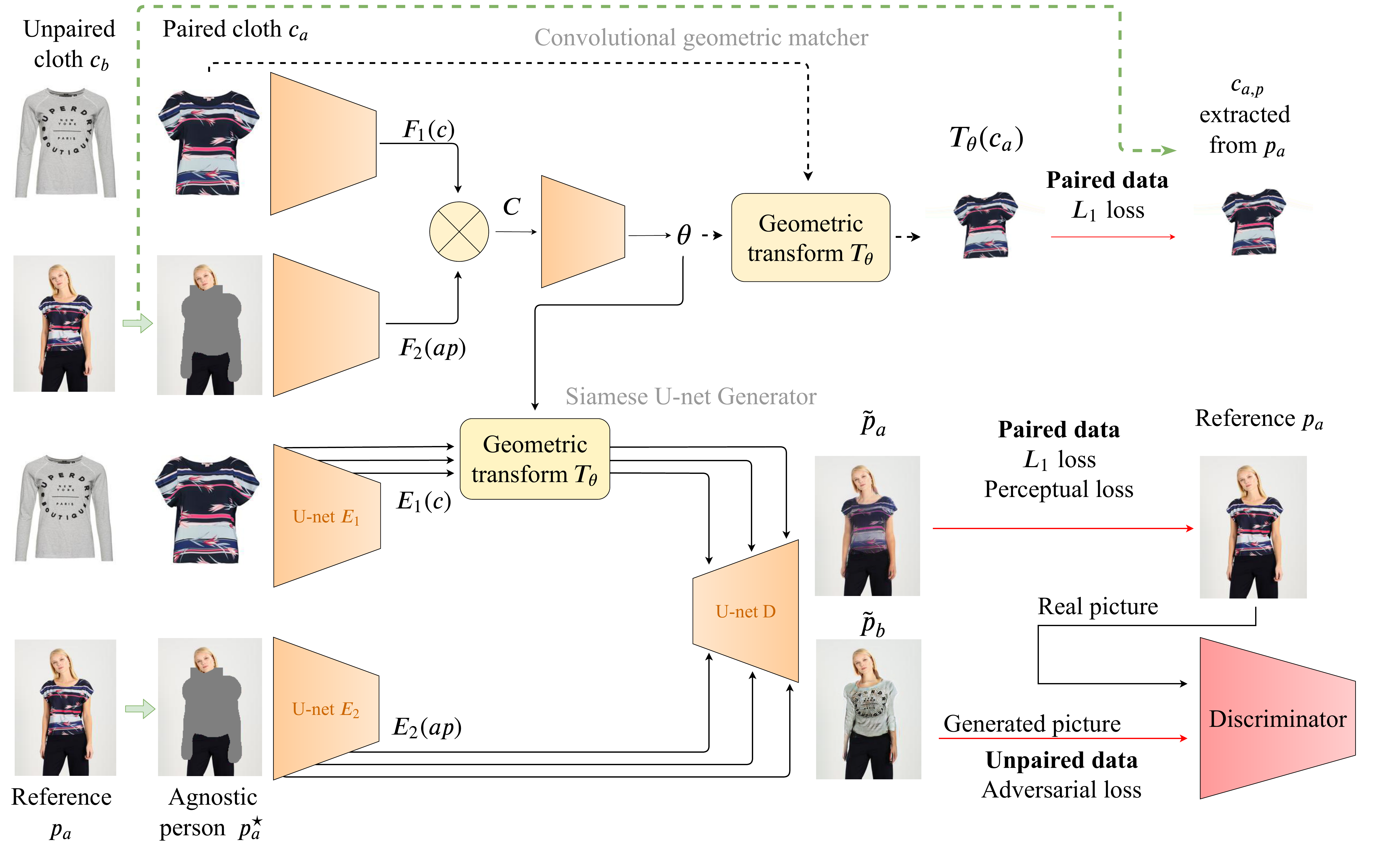}
    \caption{\label{fig:architecture} The teacher T-WUTON : our proposed end-to-end warping U-net architecture. Dotted arrows correspond to the forward pass only performed during training. Green arrows are the human parser, red ones are the loss functions. The geometric transforms share the same parameters but do not operate on the same spaces. The different training procedure for paired and unpaired pictures is explained in Section \ref{sec:training}.}
\end{figure*}
Our warping U-net is composed of two connected modules, as shown in Fig. \ref{fig:architecture}. The first one is a convolutional geometric matcher, which has a similar architecture as \cite{rocco2017convolutional,wang2018toward}. It outputs the parameters $\theta$ of a geometric transformation, a TPS transform in our case. This geometric transformation aligns the in-shop cloth image with the reference person. However, in contrast to previous work \cite{dong2019towards,han2018viton,wang2018toward}, we use the geometric transformation on the feature maps of the generator rather than at a pixel-level. Thus, we learn to deform the feature maps that pass through the skip connections of the second module, a U-net \cite{ronneberger2015u} generator which synthesizes the output image $\Tilde{p}$.

The architecture of the convolutional geometric matcher is taken from CP-VTON \cite{wang2018toward}, which reuses the generic geometric matcher from \cite{rocco2017convolutional}. It is composed of two feature extractors $F_1$ and $F_2$, which are standard convolutional neural networks. The local vectors of feature maps $F_1(c)$ and $F_2(p^\star)$ are then L2-normalized and a correlation map $C$ is computed as follows:
\begin{equation}
     C_{ijk} = F_{1_{i,j}}(c) \cdot F_{2_{m,n}}(p^\star)
\end{equation}
where k is the index for the position (m, n). This correlation map captures dependencies between distant locations of the two feature maps, which is useful to align the two images. $C$ is the input of a regression network, which outputs the parameters $\theta$ and allows to perform the geometric transformation $T_{\theta}$. We use TPS transformations \cite{bookstein1989principal}, which generate smooth sampling grids given control points. Each scale of the U-net is transformed with the same parameters $\theta$.

The input of the U-net generator is also the tuple of pictures $(p^\star,c)$. Since these two images are not spatially aligned, we cannot simply concatenate them and feed a standard U-net. To alleviate this, we use two different encoders $E_1$ and $E_2$ processing each image independently and with non-shared parameters. Then, the feature maps of the in-shop cloth $E_1(c)$ are transformed at each scale $i$: $E^i_1(c) = T_{\theta} (E^i_1(c))$. Then, the feature maps of the two encoders are concatenated and feed the decoder at each scale. With aligned feature maps, the generator is able to compose them and to produce realistic results. Feature maps warping was also proposed in \cite{dong2018soft,siarohin2018deformable}.  We use instance normalization in the U-net generator, which is more effective than batch normalization \cite{ioffe2015batch} for image generation \cite{ulyanov2017improved}.

\subsection{Training procedure of the teacher model}
\label{sec:training}
We will now detail the training procedure of T-WUTON, \textit{i.e.} the data representation and the different loss functions of the teacher model.

While previous works use a rich person representation with more than 20 channels representing human pose, body shape and the RGB image of the head, we only mask the upper-body of the reference person. Our agnostic person representation $p^\star$ is thus a 3-channel RGB image with a masked area. We compute the upper-body mask from pose and body parsing information provided by a pre-trained neural network from \cite{liang2019look}. Precisely, we mask the areas corresponding to the arms, the upper-body cloth and a bounding box around the neck.

Using the dataset from \cite{dong2019towards}, we have pairs of in-shop cloth image $c_a$ and a person wearing the same cloth $p_a$. Using a human parser and a human pose estimator, we generate $p^{\star}_a$. From the parsing information, we can also isolate the cloth on the image $p_a$ and get $c_{a,p}$, the cloth worn by the reference person. Moreover, we get the image of another in-shop cloth $c_b$. The inputs of our network are the two tuples $(p^{\star}_a, c_a)$ and $(p^{\star}_a, c_b)$. The outputs are respectively $(\Tilde{p}_a, \theta_a)$ and $(\Tilde{p}_b, \theta_b)$.

The cloth worn by the person $c_{a,p}$ allows us to guide directly the geometric matcher with a $L_1$ loss:
\begin{equation}
    L_{warp} = \lVert T_{\theta_a} (c_a) - c_{a,p} \rVert_1
\end{equation}
The image $p_a$ of the reference person provides a supervision for the whole pipeline. Similarly to CP-VTON \cite{wang2018toward}, we use two different losses to guide the generation of the final image $\Tilde{p}_a$, the pixel-level $L_1$ loss $\lVert \Tilde{p}_a - p_a \rVert_1$ and the perceptual loss \cite{johnson2016perceptual}.
We focus on $L_1$ losses since they are known to generate less blur than $L_2$ for image generation \cite{zhao2016loss}.
The latter consists of using the features extracted with a pre-trained neural network, VGG \cite{simonyan2014very} in our case. Specifically, our perceptual loss is:
\begin{equation}
    L_{perceptual} = \sum_{i = 1}^{5} \lVert \phi_i(\Tilde{p}_a) - \phi_i(p_a) \rVert_1
\end{equation}
where $\phi_i (I)$ are the feature maps of an image I extracted at the i-th layer of the VGG network.
Furthermore, we exploit adversarial training to train the network to fit $c_b$ on the same agnostic person representation $p^{\star}_a$, which is extracted from a person wearing $c_a$. This is only feasible with an adversarial loss, since there is no available ground-truth for this pair $(p^{\star}_a, c_b)$. Thus, we feed the discriminator with the synthesized image $\Tilde{p}_b$ and real images of persons from the dataset. This adversarial loss is also back-propagated to the convolutional geometric matcher, which allows to generate much more realistic spatial transformations. We use the relativistic adversarial loss \cite{jolicoeur-martineau2018} with gradient-penalty \cite{arjovsky2017wasserstein,gulrajani2017improved}, which trains the discriminator to predict relative realness of real images compared to synthesized ones.
Finally, we optimize with Adam \cite{kingma2014adam} the following objective function:
\begin{equation}
     L = \lambda_{w} L_{warp} + \lambda_{p} L_{perceptual} + \lambda_{L_1} L_1 + \lambda_{adv} L_{adv}
\end{equation}

\subsection{Training procedure of the student model}
\label{sec:self-supervision}
\begin{wrapfigure}{tr}{0.5\textwidth}
    \vspace{-25pt}
    \centering
    \includegraphics[trim={14cm 3cm 9cm 5cm}, width=\linewidth, clip]{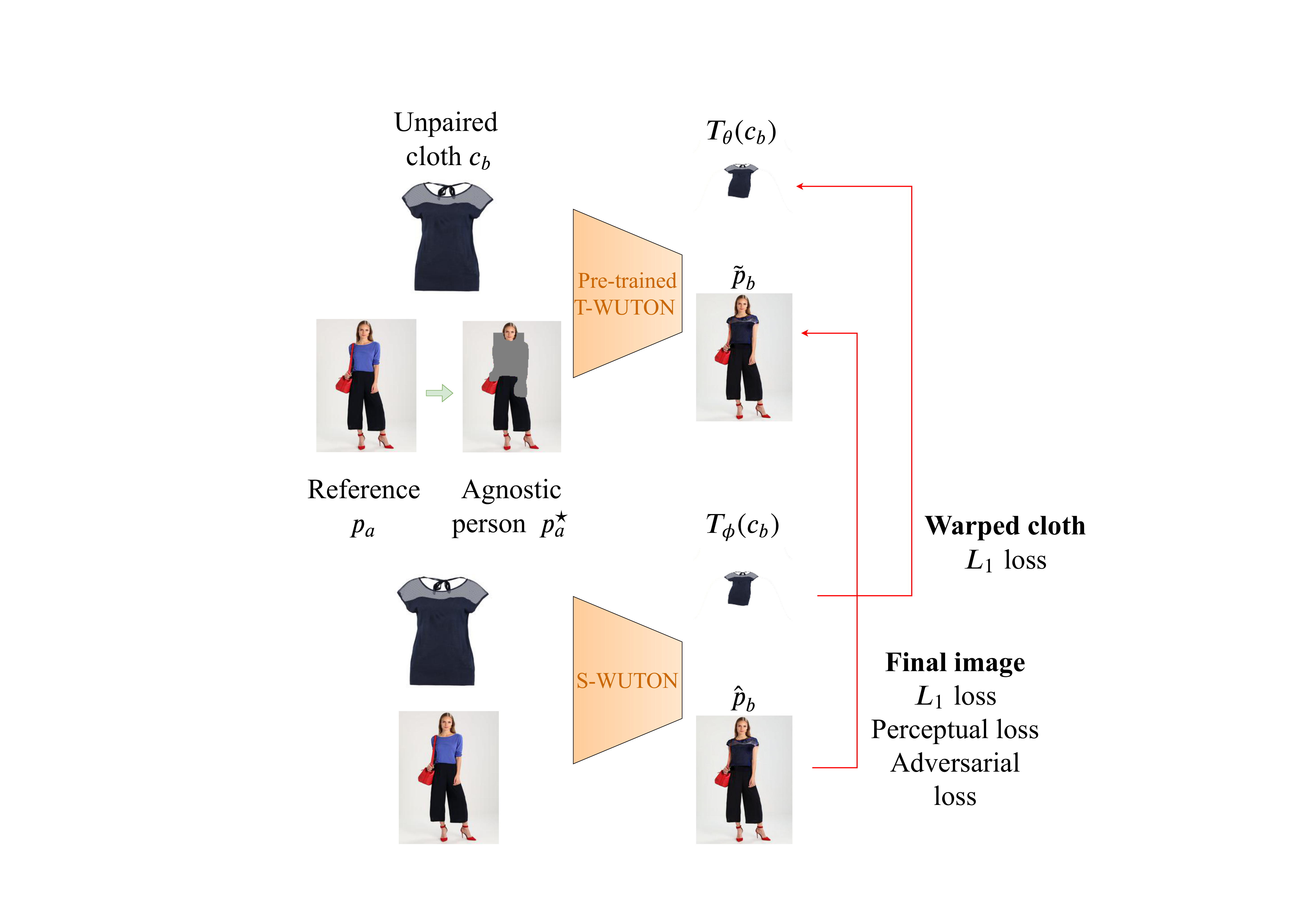}
  \caption{\label{fig:self} S-WUTON: our training scheme allowing to remove the need for a human parser at inference time. We use human parser and pre-trained T-WUTON to generate synthetic ground-truth for a student model S-WUTON.}
\end{wrapfigure}

We propose to use a teacher-student approach to distill the pipeline composed of $\{h,\text{T-WUTON}\}$ in a single student WUTON (S-WUTON). Indeed, our pre-trained T-WUTON is able to generate realistic images and geometric deformations of clothes on images pre-processed by $h$. We leverage it and use it as a way to construct generated triplets $\{(p_a,c_b),\Tilde{p}_b\}$, where $\Tilde{p}_b$ is the image synthesized by T-WUTON. With this pre-trained model, we can supervise the training of a student model S-WUTON. This allows to train the student model on the initial task of changing the cloth rather than reconstructing the upper-body. The student model has the exact same architecture than T-WUTON but different inputs and ground-truth. Hence, its inputs are $(p_a, c_b)$, where $p_a$ is the non-masked image of a person. Having this non-masked image as input, the student model does not need a human parser for pre-processing images. The ground-truth of S-WUTON are the outputs of T-WUTON, for both the warped cloth $T_\theta (c_b)$ and the final synthesized image $\Tilde{p}_b$. The training scheme of the student model S-WUTON is shown in Fig. \ref{fig:self}.

More precisely, let us define the inputs-outputs of the teacher and student model: $(\hat{p}_b, \phi) = \text{S-WUTON} (p_a,c_b)$ and $(\Tilde{p}_b, \theta) = \text{T-WUTON} (h(p_a),c_b)$. Then, the loss functions of S-WUTON are:
\begin{equation}
     L_{warp} = \lVert T_{\phi} (c_b) - T_{\theta} (c_b) \rVert_1
\end{equation}

\begin{equation}
    L_{perceptual} = \sum_{i = 1}^{5} \lVert \phi_i(\hat{p}_b) - \phi_i(\Tilde{p}_b) \rVert_1
\end{equation}

\begin{equation}
   L_1 = \lVert \hat{p}_b - \Tilde{p}_b\rVert_1
\end{equation}
Finally, the total loss of the student model is:
\begin{equation}
     L = \lambda_{w} L_{warp} + \lambda_{p} L_{perceptual} + \lambda_{L_1} L_1 + \lambda_{adv} L_{adv}
\end{equation}

The adversarial loss $L_{adv}$ is independant from T-WUTON. Here, we also use the relativistic loss with gradient penalty on the discriminator. The real data consists of images of persons from the dataset $p_a$, and the fake data corresponds to the synthesized images $\hat{p}_b$. Notice that without the adversarial loss, it would be a standard teacher-student setting, where the student is only guided by the outputs of the teacher. In our case, the discriminator (\textit{i.e.} $L_{adv}$) helps S-WUTON to be close to the real data distribution, and not only to the teacher's distribution. As shown by the ablation study in Section \ref{sec:ablation_distillation}, it is an important component and is necessary to make S-WUTON exploit the components that are masked from T-WUTON (\textit{e.g.} hands).

\section{Experiments and analysis}

\addtolength{\tabcolsep}{-6pt}
\begin{figure*}[ht]
\fontsize{7}{7}\selectfont
\centering
\begin{tabular*}{340pt}{@{\extracolsep{\fill}}cccccccccc}
Ref. & Target & CP- & T-WUT- & S-WUTON & Ref. & Target & CP- & VTNFP & S-WUTON\\
person & cloth & VTON & ON(ours) & (ours) & person & cloth & VTON & & (ours) \\
\subfigure{\includegraphics[width=0.099\linewidth]{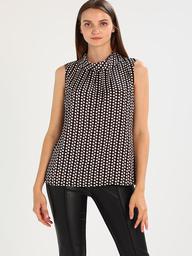}}  &
\subfigure{\includegraphics[width=0.099\linewidth]{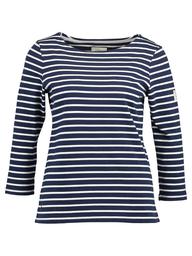}} &
\subfigure{\includegraphics[width=0.099\linewidth]{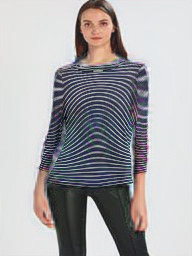}} &
\subfigure{\includegraphics[width=0.099\linewidth]{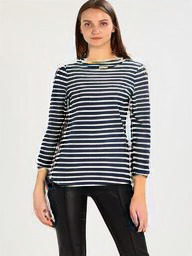}} &
\subfigure{\includegraphics[width=0.099\linewidth]{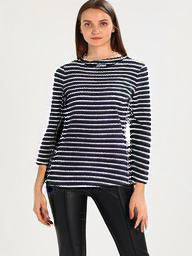}} &
\subfigure{\includegraphics[width=0.099\linewidth]{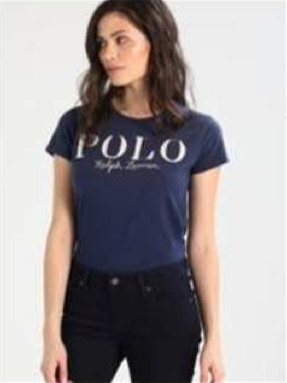}} &
\subfigure{\includegraphics[width=0.099\linewidth]{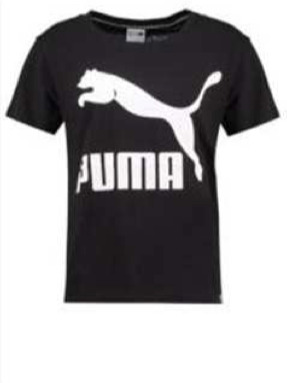}} &
\subfigure{\includegraphics[width=0.099\linewidth]{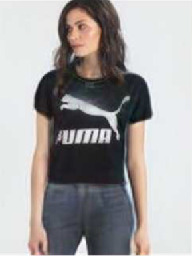}} &
\subfigure{\includegraphics[width=0.099\linewidth]{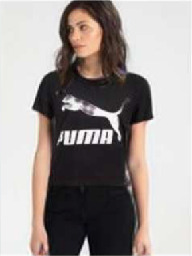}} &
\subfigure{\includegraphics[width=0.099\linewidth]{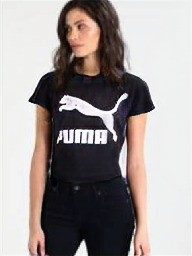}} \\ [-2.2ex]
\subfigure{\includegraphics[width=0.099\linewidth]{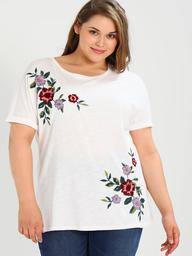}} &
\subfigure{\includegraphics[width=0.099\linewidth]{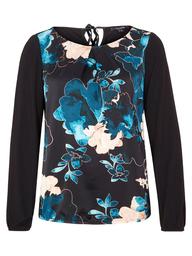}} &
\subfigure{\includegraphics[width=0.099\linewidth]{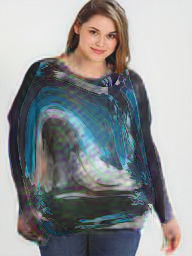}} &
\subfigure{\includegraphics[width=0.099\linewidth]{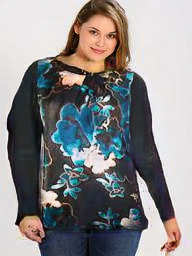}} &
\subfigure{\includegraphics[width=0.099\linewidth]{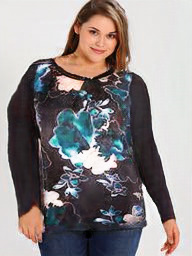}} &
\subfigure{\includegraphics[width=0.099\linewidth]{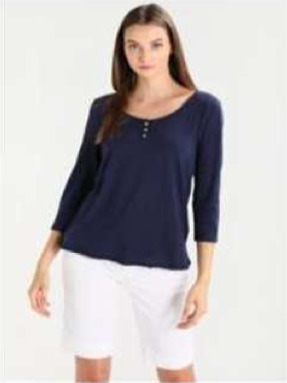}} &
\subfigure{\includegraphics[width=0.099\linewidth]{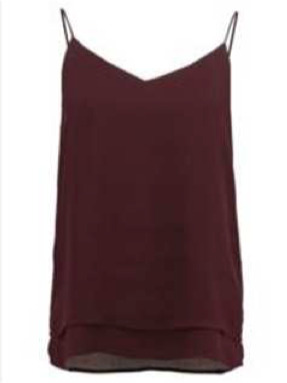}} &
\subfigure{\includegraphics[width=0.099\linewidth]{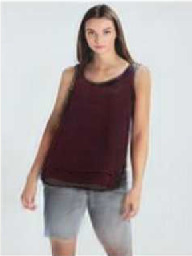}} &
\subfigure{\includegraphics[width=0.099\linewidth]{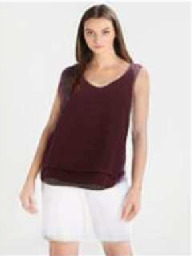}} &
\subfigure{\includegraphics[width=0.099\linewidth]{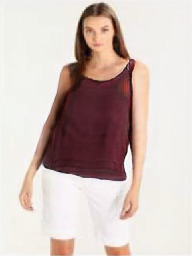}} \\ [-2.2ex]
\subfigure{\includegraphics[width=0.099\linewidth]{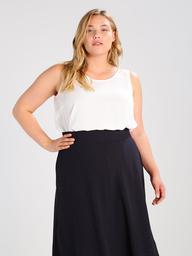}} &
\subfigure{\includegraphics[width=0.099\linewidth]{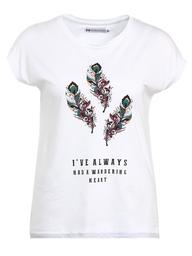}} &
\subfigure{\includegraphics[width=0.099\linewidth]{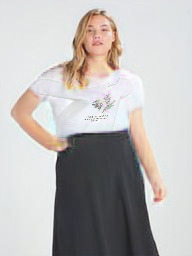}} &
\subfigure{\includegraphics[width=0.099\linewidth]{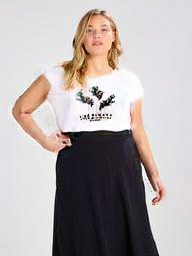}} &
\subfigure{\includegraphics[width=0.099\linewidth]{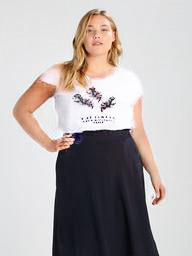}} &
\subfigure{\includegraphics[width=0.099\linewidth]{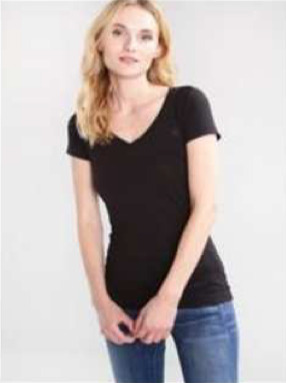}} &
\subfigure{\includegraphics[width=0.099\linewidth]{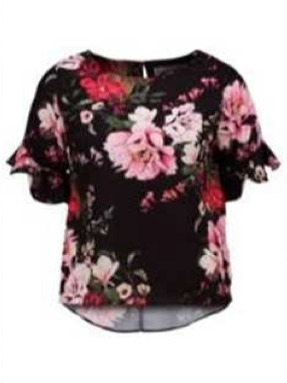}} &
\subfigure{\includegraphics[width=0.099\linewidth]{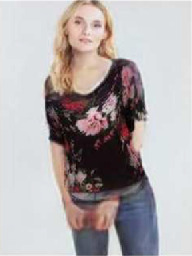}} &
\subfigure{\includegraphics[width=0.099\linewidth]{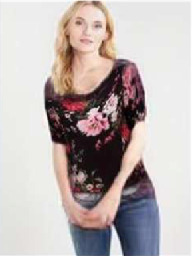}} &
\subfigure{\includegraphics[width=0.099\linewidth]{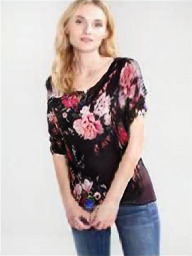}}\\ [-2.2ex]
 \subfigure{\includegraphics[width=0.099\linewidth]{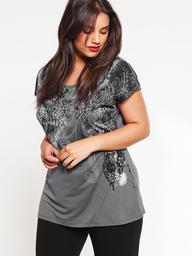}} &
\subfigure{\includegraphics[width=0.099\linewidth]{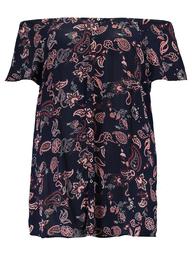}} &
\subfigure{\includegraphics[width=0.099\linewidth]{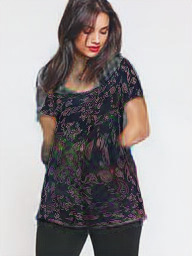}} &
\subfigure{\includegraphics[width=0.099\linewidth]{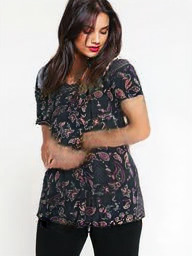}} &
\subfigure{\includegraphics[width=0.099\linewidth]{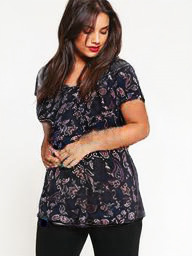}} &
\subfigure{\includegraphics[width=0.099\linewidth]{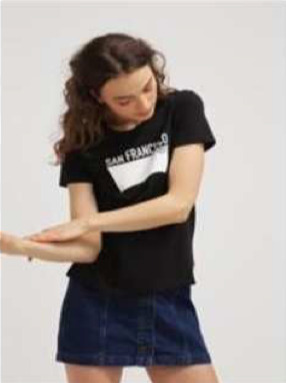}} &
\subfigure{\includegraphics[width=0.099\linewidth]{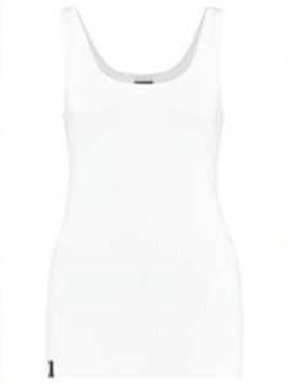}} &
\subfigure{\includegraphics[width=0.099\linewidth]{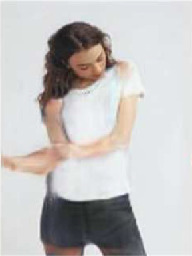}} &
\subfigure{\includegraphics[width=0.099\linewidth]{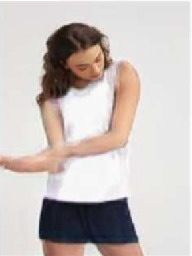}} &
\subfigure{\includegraphics[width=0.099\linewidth]{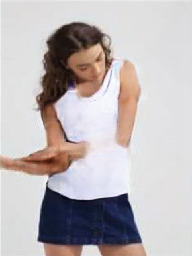}}\\
\end{tabular*}
\caption{\label{fig:comparison} On the left side, comparison of our method with CP-VTON \cite{wang2018toward}. For fairness, the two methods are trained on the same dataset and on the same agnostic person representation $p^\star$. More examples are provided in supplementary material. On the right side, comparison with recent work VTNFP. Except for S-WUTON's column, images are taken from their paper. }
\end{figure*}
\addtolength{\tabcolsep}{6pt}

We first describe the dataset. We then compare our approach with CP-VTON
\cite{wang2018toward}, a current state-of-the-art for the virtual try-on task. We present visual and quantitative results proving that S-WUTON achieves state-of-the-art results, and that the distillation process allows to improve image quality. We show that this stands for several metrics, and with a user study. We then provide a comparison of the runtime of virtual try-on algorithms on a Tesla NVIDIA V100 GPU. The teacher-student distillation allows to decrease the runtime by an order of magnitude. Finally, we outline the importance of the adversarial loss in our teacher-student setting.

We also show some visual comparisons with recent work VTNFP \cite{Yu_2019_ICCV}. Images are taken from their paper. However, since their model is not available, we could not compute the other metrics. We provide more visual comparisons with VTNFP and ClothFlow \cite{Han_2019_ICCV} in supplementary material.

\subsection{Dataset}
For copyright issues, we can not use the dataset from VITON \cite{han2018viton} and CP-VTON \cite{wang2018toward}. Instead, we leverage the \textit{Image-based Multi-pose Virtual try-on} dataset.
This dataset contains 35,687/13,524 person/cloth images at 256x192 resolution. 4175 pairs are kept for test so the cloth was not seen during training. A random shuffle of these pairs produces the unpaired person/cloth images.  For each in-shop cloth image, there are multiple images of a model wearing the given cloth from different views and in different poses. We remove images tagged as back images since the in-shop cloth image is only from the front. We process the images with a neural human parser and pose estimator, specifically the joint body parsing and pose estimation network \cite{gong2017look,liang2019look}.

\subsection{Visual results}
\begin{figure*}[]
\fontsize{8}{8}\selectfont
\begin{tabular*}{340pt}{@{\extracolsep{\fill}}ccccccccccc}
 &
\subfigure{\includegraphics[width=0.10\linewidth]{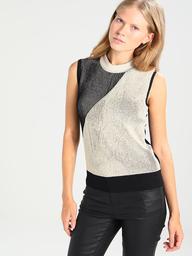}} &
\subfigure{\includegraphics[width=0.10\linewidth]{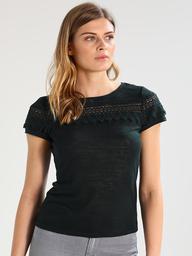}} &
\subfigure{\includegraphics[width=0.10\linewidth]{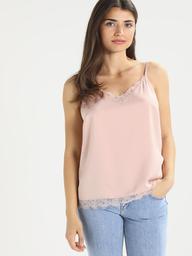}} &
\subfigure{\includegraphics[width=0.10\linewidth]{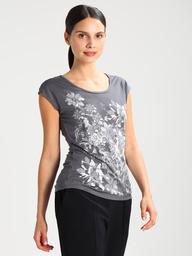}} &
\subfigure{\includegraphics[width=0.10\linewidth]{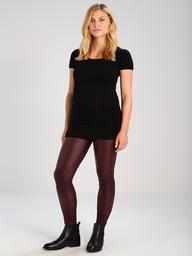}} &
\subfigure{\includegraphics[width=0.10\linewidth]{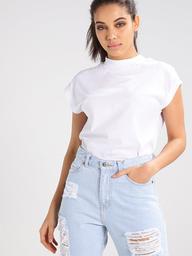}} &
\subfigure{\includegraphics[width=0.10\linewidth]{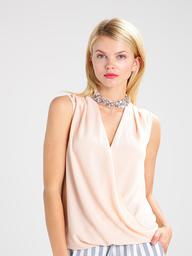}} &
\subfigure{\includegraphics[width=0.10\linewidth]{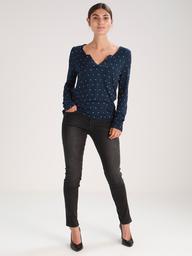}} &
\subfigure{\includegraphics[width=0.10\linewidth]{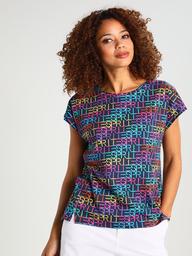}}\\ [-2.2ex]
\subfigure{\includegraphics[width=0.10\linewidth]{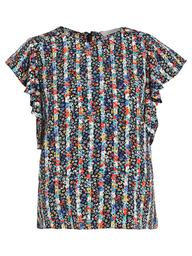}} &
\subfigure{\includegraphics[width=0.10\linewidth]{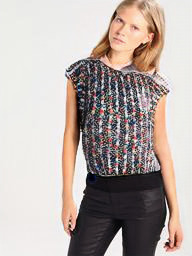}} &
\subfigure{\includegraphics[width=0.10\linewidth]{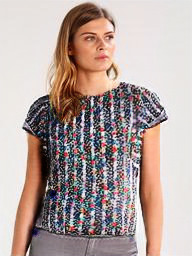}} &
\subfigure{\includegraphics[width=0.10\linewidth]{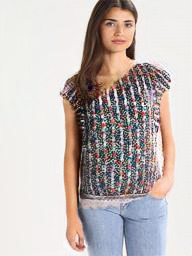}} &
\subfigure{\includegraphics[width=0.10\linewidth]{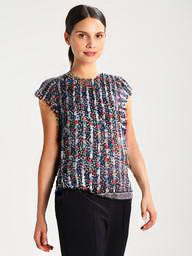}} &
\subfigure{\includegraphics[width=0.10\linewidth]{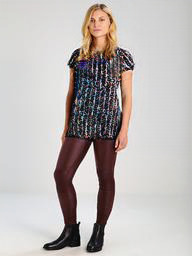}} &
\subfigure{\includegraphics[width=0.10\linewidth]{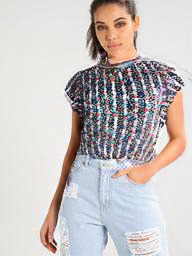}} &
\subfigure{\includegraphics[width=0.10\linewidth]{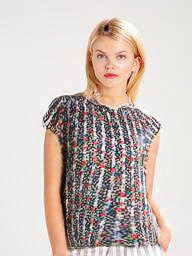}} &
\subfigure{\includegraphics[width=0.10\linewidth]{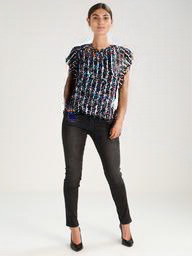}} &
\subfigure{\includegraphics[width=0.10\linewidth]{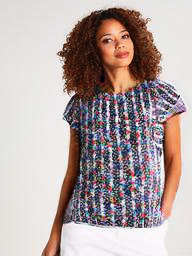}} \\ [-2.2ex]
\subfigure{\includegraphics[width=0.10\linewidth]{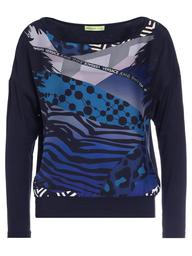}} &
\subfigure{\includegraphics[width=0.10\linewidth]{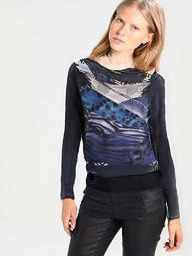}} &
\subfigure{\includegraphics[width=0.10\linewidth]{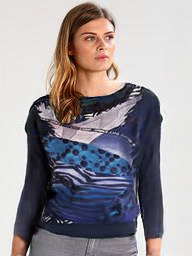}} &
\subfigure{\includegraphics[width=0.10\linewidth]{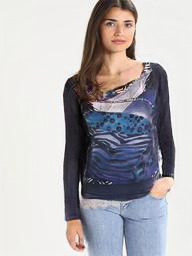}} &
\subfigure{\includegraphics[width=0.10\linewidth]{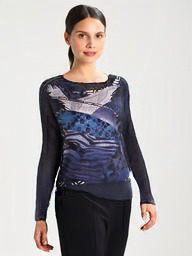}} &
\subfigure{\includegraphics[width=0.10\linewidth]{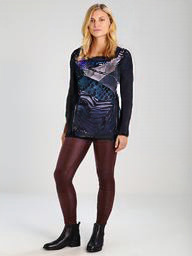}} &
\subfigure{\includegraphics[width=0.10\linewidth]{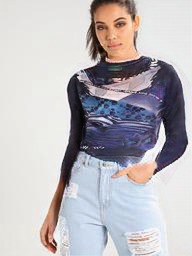}} &
\subfigure{\includegraphics[width=0.10\linewidth]{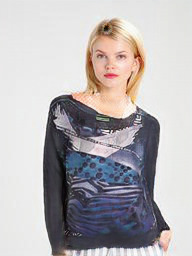}} &
\subfigure{\includegraphics[width=0.10\linewidth]{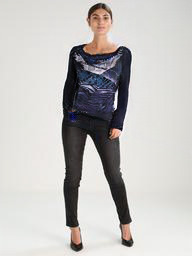}} &
\subfigure{\includegraphics[width=0.10\linewidth]{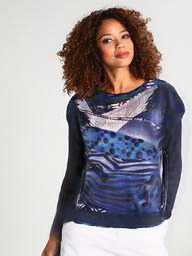}} \\ [-2.2ex]
\subfigure{\includegraphics[width=0.10\linewidth]{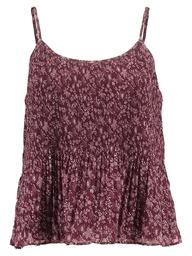}} &
\subfigure{\includegraphics[width=0.10\linewidth]{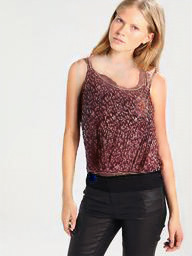}} &
\subfigure{\includegraphics[width=0.10\linewidth]{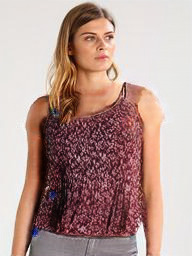}} &
\subfigure{\includegraphics[width=0.10\linewidth]{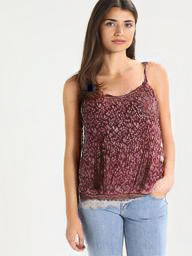}} &
\subfigure{\includegraphics[width=0.10\linewidth]{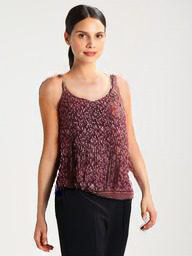}} &
\subfigure{\includegraphics[width=0.10\linewidth]{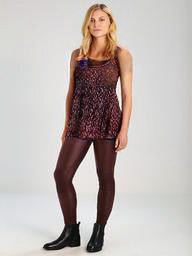}} &
\subfigure{\includegraphics[width=0.10\linewidth]{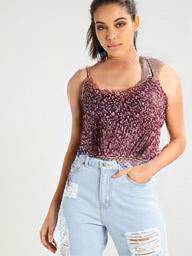}} &
\subfigure{\includegraphics[width=0.10\linewidth]{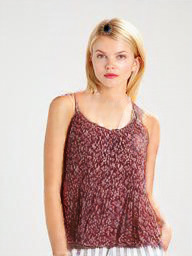}} &
\subfigure{\includegraphics[width=0.10\linewidth]{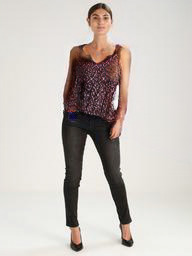}} &
\subfigure{\includegraphics[width=0.10\linewidth]{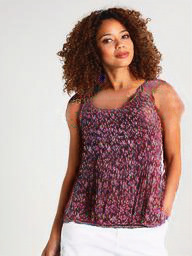}} \\ [-2.2ex]
\subfigure{\includegraphics[width=0.10\linewidth]{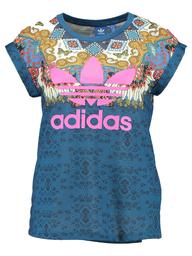}} &
\subfigure{\includegraphics[width=0.10\linewidth]{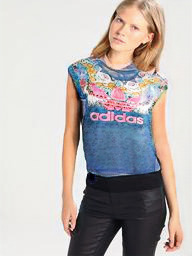}} &
\subfigure{\includegraphics[width=0.10\linewidth]{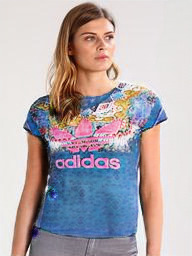}} &
\subfigure{\includegraphics[width=0.10\linewidth]{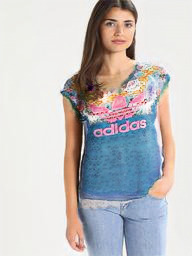}} &
\subfigure{\includegraphics[width=0.10\linewidth]{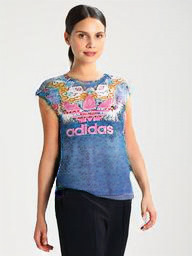}} &
\subfigure{\includegraphics[width=0.10\linewidth]{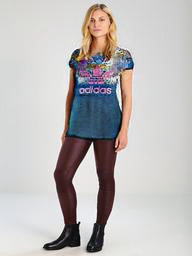}} &
\subfigure{\includegraphics[width=0.10\linewidth]{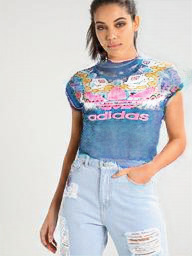}} &
\subfigure{\includegraphics[width=0.10\linewidth]{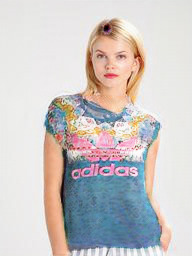}} &
\subfigure{\includegraphics[width=0.10\linewidth]{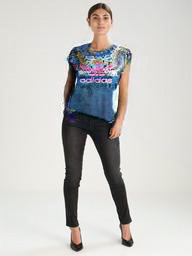}} &
\subfigure{\includegraphics[width=0.10\linewidth]{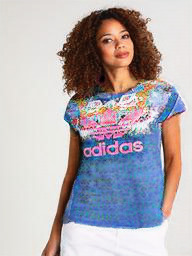}} \\ [-2.2ex]
\subfigure{\includegraphics[width=0.10\linewidth]{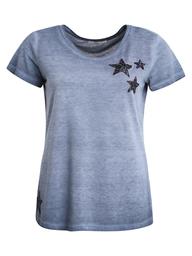}} &
\subfigure{\includegraphics[width=0.10\linewidth]{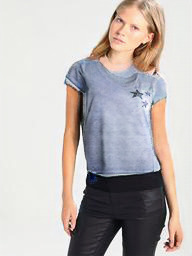}} &
\subfigure{\includegraphics[width=0.10\linewidth]{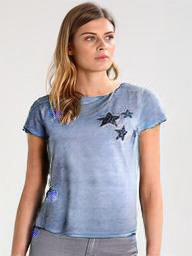}} &
\subfigure{\includegraphics[width=0.10\linewidth]{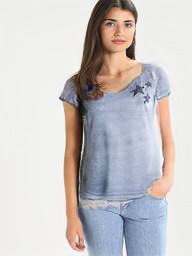}} &
\subfigure{\includegraphics[width=0.10\linewidth]{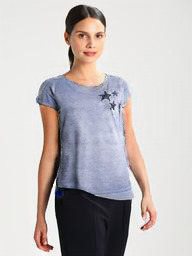}} &
\subfigure{\includegraphics[width=0.10\linewidth]{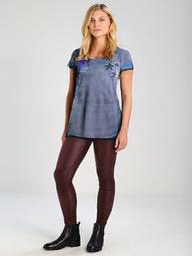}} &
\subfigure{\includegraphics[width=0.10\linewidth]{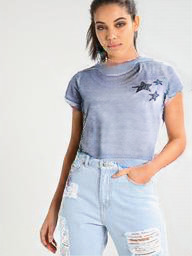}} &
\subfigure{\includegraphics[width=0.10\linewidth]{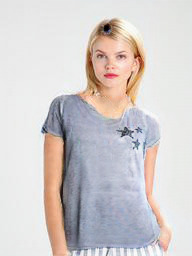}} &
\subfigure{\includegraphics[width=0.10\linewidth]{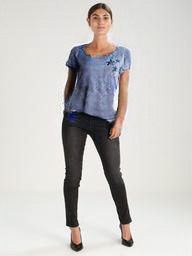}} &
\subfigure{\includegraphics[width=0.10\linewidth]{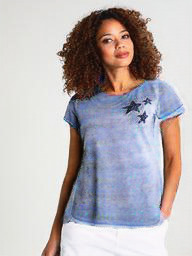}} \\ [-2.2ex]
\subfigure{\includegraphics[width=0.10\linewidth]{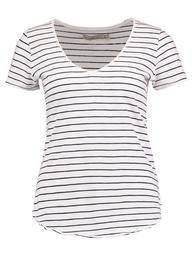}} &
\subfigure{\includegraphics[width=0.10\linewidth]{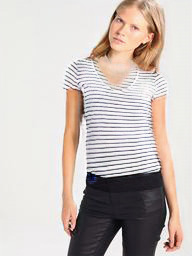}} &
\subfigure{\includegraphics[width=0.10\linewidth]{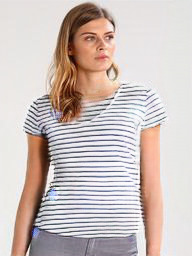}} &
\subfigure{\includegraphics[width=0.10\linewidth]{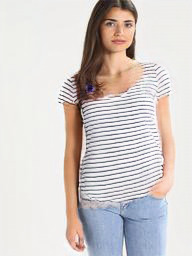}} &
\subfigure{\includegraphics[width=0.10\linewidth]{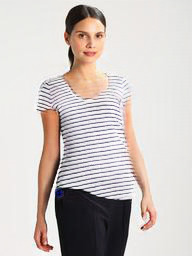}} &
\subfigure{\includegraphics[width=0.10\linewidth]{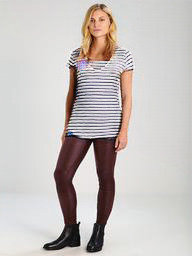}} &
\subfigure{\includegraphics[width=0.10\linewidth]{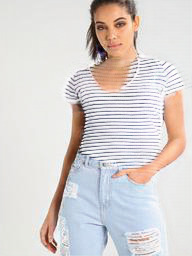}} &
\subfigure{\includegraphics[width=0.10\linewidth]{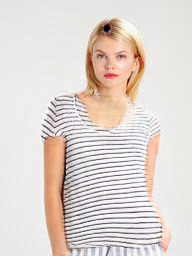}} &
\subfigure{\includegraphics[width=0.10\linewidth]{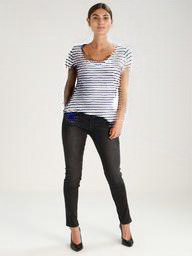}} &
\subfigure{\includegraphics[width=0.10\linewidth]{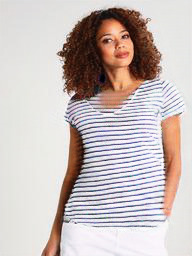}} \\ [-2.2ex]
\subfigure{\includegraphics[width=0.10\linewidth]{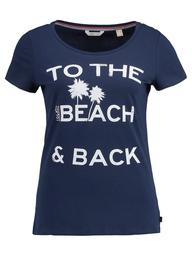}} &
\subfigure{\includegraphics[width=0.10\linewidth]{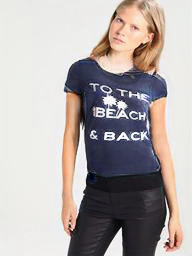}} &
\subfigure{\includegraphics[width=0.10\linewidth]{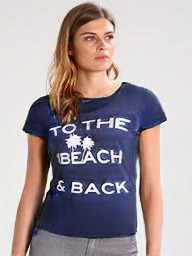}} &
\subfigure{\includegraphics[width=0.10\linewidth]{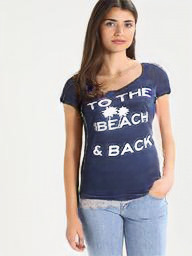}} &
\subfigure{\includegraphics[width=0.10\linewidth]{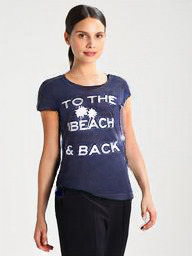}} &
\subfigure{\includegraphics[width=0.10\linewidth]{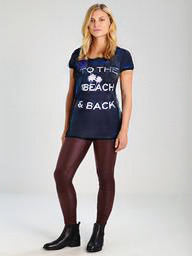}} &
\subfigure{\includegraphics[width=0.10\linewidth]{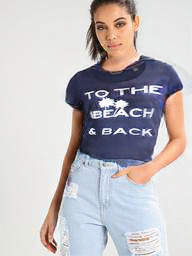}} &
\subfigure{\includegraphics[width=0.10\linewidth]{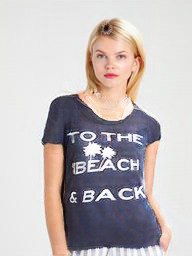}} &
\subfigure{\includegraphics[width=0.10\linewidth]{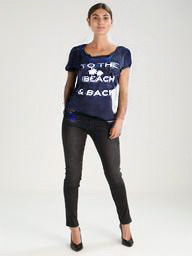}} &
\subfigure{\includegraphics[width=0.10\linewidth]{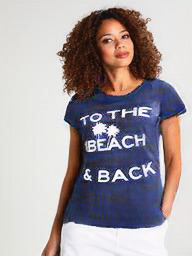}} \\ [-2.2ex]
\subfigure{\includegraphics[width=0.10\linewidth]{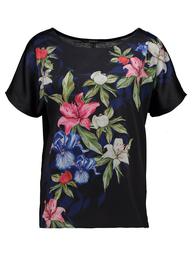}} &
\subfigure{\includegraphics[width=0.10\linewidth]{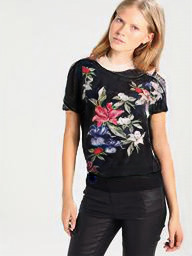}} &
\subfigure{\includegraphics[width=0.10\linewidth]{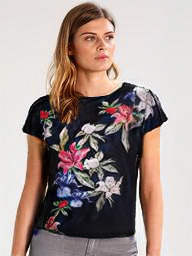}} &
\subfigure{\includegraphics[width=0.10\linewidth]{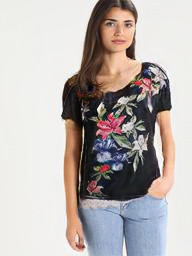}} &
\subfigure{\includegraphics[width=0.10\linewidth]{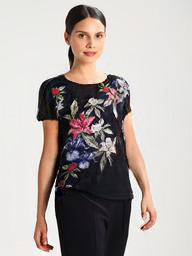}} &
\subfigure{\includegraphics[width=0.10\linewidth]{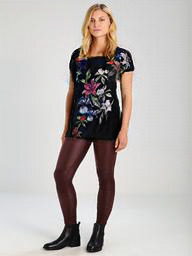}} &
\subfigure{\includegraphics[width=0.10\linewidth]{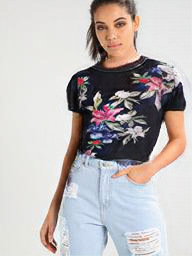}} &
\subfigure{\includegraphics[width=0.10\linewidth]{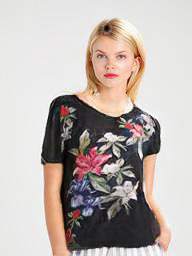}} &
\subfigure{\includegraphics[width=0.10\linewidth]{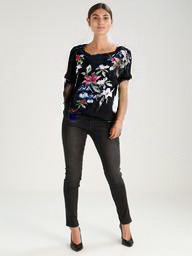}} &
\subfigure{\includegraphics[width=0.10\linewidth]{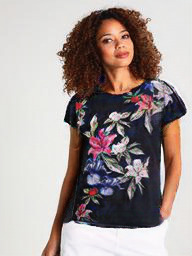}} \\ [-2.2ex]
\subfigure{\includegraphics[width=0.10\linewidth]{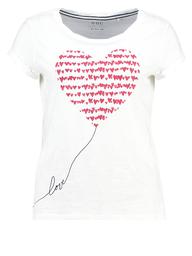}} &
\subfigure{\includegraphics[width=0.10\linewidth]{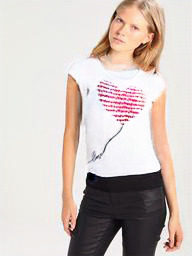}} &
\subfigure{\includegraphics[width=0.10\linewidth]{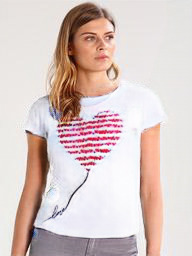}} &
\subfigure{\includegraphics[width=0.10\linewidth]{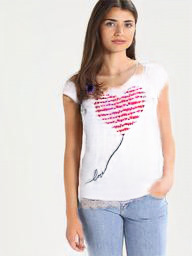}} &
\subfigure{\includegraphics[width=0.10\linewidth]{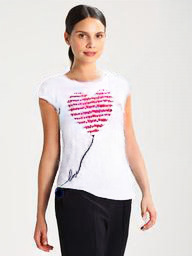}} &
\subfigure{\includegraphics[width=0.10\linewidth]{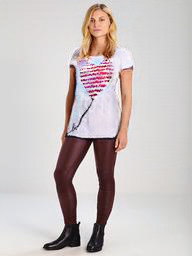}} &
\subfigure{\includegraphics[width=0.10\linewidth]{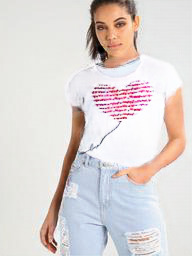}} &
\subfigure{\includegraphics[width=0.10\linewidth]{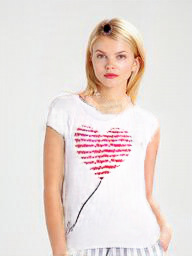}} &
\subfigure{\includegraphics[width=0.10\linewidth]{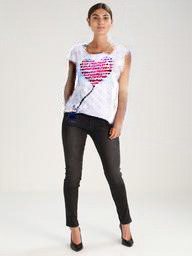}} &
\subfigure{\includegraphics[width=0.10\linewidth]{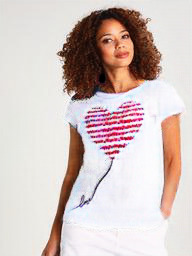}} \\ [-2.2ex]
\end{tabular*}
\caption{\label{fig:cross_examples} Our student model S-WUTON generates high-quality images and preserves both person's and cloth's attributes.}
\end{figure*}

Visual results of our method and CP-VTON are shown in Fig. \ref{fig:comparison}. On the left side, images are computed from models trained on MG-VTON dataset, with $p^\star_{t-wuton}$ representation for T-WUTON and CP-VTON for fairness. On the right side, images are taken from VTNFP paper \cite{Yu_2019_ICCV}. There, CP-VTON and VTNFP were trained on the original dataset from VITON, and CP-VTON uses $p^\star_{cp-vton}$. More images from S-WUTON are provided in Fig. \ref{fig:introduction} and Fig. \ref{fig:cross_examples}.

CP-VTON has trouble to realistically deform and render complex patterns like stripes or flowers. Control points of the $T_\theta$ transform are visible and lead to unrealistic curves and deformations on the clothes. Also, the edges of cloth patterns and body contours are blurred.

Firstly, our proposed T-WUTON architecture allows to improve the baseline CP-VTON. Indeed, our method generates spatial transformations of a much higher visual quality, which is specifically visible for stripes (1st row). It is able to preserve complex visual patterns of clothes and produces sharper images than CP-VTON and VTNFP on the edges. Secondly, we can observe the importance of our distillation process with the visual results from S-WUTON. Since it has a non-masked image as input, it is able to preserve body details, especially the hands. Moreover, as shown in Fig. \ref{fig:introduction}, S-WUTON is robust to a bad parsing and preserves a person's attributes that are important for the virtual try-on task.

Generally, our method generates results of high visual quality while preserving the characteristics of the target cloth and of the person. However, VTNFP can surpass S-WUTON when models are crossing arms (4th row, right side), which is sometimes a failure case of our method. Note that this is not general, since on  (3rd row, right side) and (4th row, left side) in Fig. \ref{fig:comparison} and on the two last columns in Fig. \ref{fig:vis_distillation}, models are crossing arms and S-WUTON manages to nicely compose the arms with the occluded cloth.

\subsection{Quantitative results}
\begin{table*}[h]
\begin{center}
\begin{tabular}{l|c|c|c|c}
Method &  LPIPS & SSIM & IS & FID \\
\hline
Real data & 0 & 1 & 3.135
                   & 0 \\
\hline
CP-VTON on $p^{\star}_{cp-vton}$  & 0.182 $\pm$ 0.049
& 0.679 $\pm$ 0.073
& 2.684
& 37.237 \\
CP-VTON on $p^{\star}_{t-wuton}$ & 0.131   $\pm$ 0.058
& 0.773 $\pm$ 0.088
& 2.938
& 16.843\\
T-WUTON & \textbf{0.101 $\pm$ 0.047}& \textbf{0.799 $\pm$ 0.089}
& 3.114
& 9.877\\
S-WUTON & NA & NA & \textbf{3.154}
& \textbf{7.927}\\ \hline
VTNFP$^\star$ & NA & 0.803 & 2.784
& NA\\
ACGPN$^\star$ & NA & 0.845 & 2.829
& NA \\

\end{tabular}
\end{center}
\caption{\label{table:lpips} Quantitative results on paired setting (LPIPS and SSIM) and on unpaired setting (IS and FID). For LPIPS and FID, the lower is the better. For SSIM and IS, the higher is the better. $\pm$ reports std. dev. The two last lines (methods with $^\star$) are the results presented in ACGPN \cite{yang2020towards}. However, it has to be taken carefully since the experiments are performed on another dataset.}
\end{table*}

To further evaluate our method, we use four different metrics. Two are designed for the paired setting, that does not allow us to evaluate S-WUTON (because the input image is not masked), and one is for the unpaired setting.  The first one for the paired setting is the linear perceptual image patch similarity (LPIPS) developed in \cite{zhang2018unreasonable}, a state-of-the-art metric for comparing pairs of images. It is very similar to the perceptual loss we use in training (see Section \ref{sec:training}) since the idea is to use the feature maps extracted by a pre-trained neural network to quantify the perceptual difference between two images. Different from the basic perceptual loss, they first unit-normalize each layer in the channel dimension and then learn a rescaling that match human perception.

Such as previous works, we also use the structural similarity (SSIM) \cite{wang2004image} in the paired setting, inception score (IS) \cite{salimans2016improved} and Fr\'echet Inception Distance (FID) \cite{heusel2017gans} in the unpaired setting. We evaluate CP-VTON \cite{wang2018toward} on their agnostic person representation $p^{\star}_{cp-vton}$ (20 channels with RGB image of head and shape/pose information) and on $p^{\star}_{t-wuton}$. Results are reported in Table \ref{table:lpips}.

\subsection{User study}
We perform A/B tests on 7 users. Each one has to vote 100 times between CP-VTON and S-WUTON synthesized images, given reference person and target cloth. The user is asked to choose for the most realistic image, that preserves both person and target cloth details. The selected 100 images are a random subset of the test set in the unpaired setting. This subset is sampled for each user and is thus different for each user. There is no time limit for the users.

Let us denote $p$ the probability that an image from S-WUTON is preferable to an image from CP-VTON. The users choose our method 88\% of the time. In terms of statistical significance, it means that we can say $p>0.85$ with a confidence level of 98.7\%.

ClothFlow and VTNFP also performed user studies where they compare to CP-VTON. The authors respectively report that users prefer their method 81.2\% and 77.4\% of the time. Note that the experiment was not performed in the same setting (dataset, number of users, number of pictures per user).

\subsection{Runtime analysis}
\begin{table*}[ht]
\begin{center}
\begin{tabular}{l|c|c|c|c|c}
& CP-VTON & VTNFP & ClothFlow & T-WUTON & S-WUTON \\ \hline
Parsing + pose & 168ms & 168ms & 168ms & 168ms & \textbf{0ms} \\ \hline
Try-on &  \textbf{9ms} & $>$9ms & $>$0ms & 13ms & 13ms \\ \hline
Total & 177ms & $>$177ms & $>$168ms  & 181ms &  \textbf{13ms} \\ \hline
\end{tabular}
\end{center}
\caption{\label{table:computational_complexity} Comparison of runtime of state-of-the-art architectures for virtual try-on. The time is computed on a NVIDIA Tesla V100 GPU.}
\end{table*}
In Table \ref{table:computational_complexity}, we compare the runtime of our method to CP-VTON, ClothFlow and VTNFP. Note that the running times are estimated on a NVIDIA V100 GPU. For the human parsing and pose estimation networks, we use state-of-the-art models from \cite{gong2017look,liang2019look}. These are based on shared neural backbones for the two tasks, which accelerates the computations.

The try-on architecture of T-WUTON and S-WUTON is slightly slower than that of CP-VTON, due to the non-shared encoder and the warping at each scale of the U-Net. However, with S-WUTON we remove the wall clock bottleneck of virtual try-on system, which is the human parsing and pose estimation. Doing so, we decrease by an order of magnitude the runtime of virtual try-on algorithms, from 6FPS to 77FPS.

We include comparisons with VTNFP and ClothFlow in the Table \ref{table:computational_complexity}. Indeed, both models use human parsing and pose estimation. For VTNFP, they add a module on top of CP-VTON architecture, so their try-on architecture takes at least 9ms per image. For ClothFlow, the use of human parser and pose estimator gives a lower bound on the total runtime.

\subsection{The impact of adversarial loss in the teacher-student setting}
\label{sec:ablation_distillation}
  \begin{minipage}[t]{\textwidth}
  \begin{minipage}[b]{1\textwidth}
  \fontsize{8}{8}\selectfont
    \centering
    \begin{tabular}{cccccccc}
    Reference & \includegraphics[valign=m,width=0.10\linewidth]{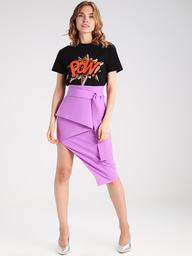} & \includegraphics[valign=m,width=0.10\linewidth]{3622_person.jpg} & {\includegraphics[valign=m,width=0.10\linewidth]{460_person.jpg}} & {\includegraphics[valign=m,width=0.10\linewidth]{717_person.jpg}} & {\includegraphics[valign=m,width=0.10\linewidth]{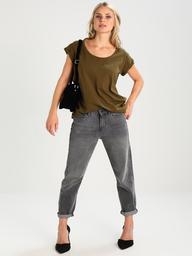}} &
    {\includegraphics[valign=m,width=0.10\linewidth]{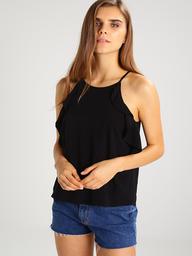}} &
    {\includegraphics[valign=m,width=0.10\linewidth]{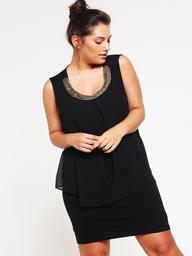}}\\
    Target cloth & \includegraphics[valign=m,width=0.10\linewidth]{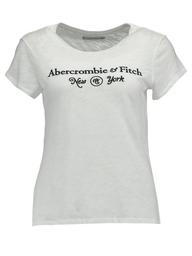} &
    \includegraphics[valign=m,width=0.10\linewidth]{3622_cloth.jpg} & {\includegraphics[valign=m,width=0.10\linewidth]{460_cloth.jpg}} & {\includegraphics[valign=m,width=0.10\linewidth]{717_cloth.jpg}} & {\includegraphics[valign=m,width=0.10\linewidth]{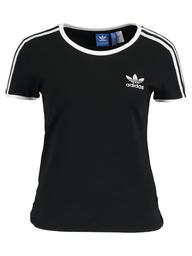}} &
    {\includegraphics[valign=m,width=0.10\linewidth]{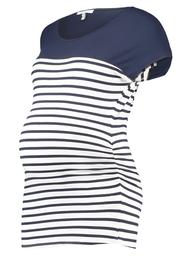}} &
    {\includegraphics[valign=m,width=0.10\linewidth]{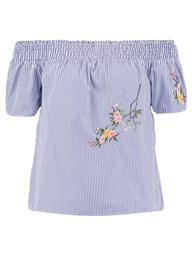}}\\
    S-WUTON w/o adv. & \includegraphics[valign=m,width=0.10\linewidth]{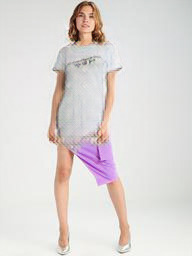} &
    \includegraphics[valign=m,width=0.10\linewidth]{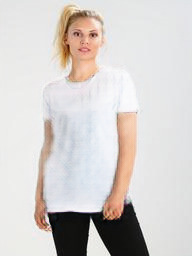} &
    {\includegraphics[valign=m,width=0.10\linewidth]{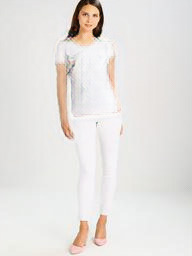}} & {\includegraphics[valign=m,width=0.10\linewidth]{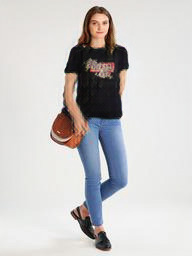}} & {\includegraphics[valign=m,width=0.10\linewidth]{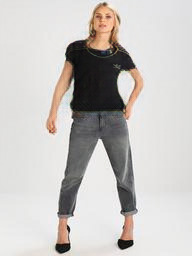}}&
    {\includegraphics[valign=m,width=0.10\linewidth]{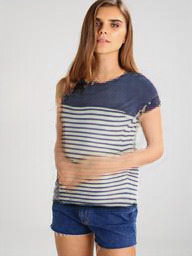}} &
    {\includegraphics[valign=m,width=0.10\linewidth]{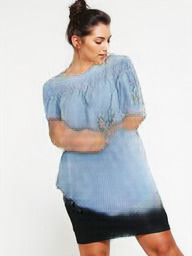}}\\
    S-WUTON & \includegraphics[valign=m,width=0.10\linewidth]{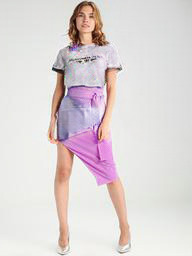} &
     \includegraphics[valign=m,width=0.10\linewidth]{3622_s_wuton.jpg} &\includegraphics[valign=m,width=0.10\linewidth]{460_S_WUTON.jpg} & \includegraphics[valign=m,width=0.10\linewidth]{717_s_wuton.jpg} & \includegraphics[valign=m,width=0.10\linewidth]{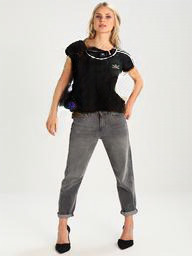}&
     \includegraphics[valign=m,width=0.10\linewidth]{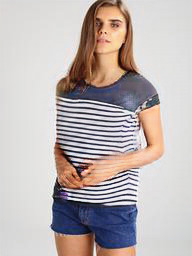} &
     \includegraphics[valign=m,width=0.10\linewidth]{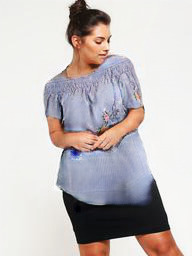}\\
    \end{tabular}
    \captionof{figure}{Visual comparison of the student model with and without the adversarial loss. Interestingly, the student model without the adversarial loss can not exploit information that is masked to the teacher, \textit{e.g.} arms and hands.\label{fig:vis_distillation}}
  \end{minipage}
 \end{minipage} \\ \\

 \begin{wraptable}{r}{5.5cm}
\begin{tabular}{c|c|c}
      & S-WUTON & S-WUTON  \\
       &  w/o adv. & \\ \hline
      IS & 2.912
      &  \textbf{3.154}
      \\
      FID & 12.620 &  \textbf{7.927}  \\
      \end{tabular}
      \caption{Comparison of IS and FID scores of S-WUTON and S-WUTON without the adversarial loss.}\label{wrap-tab:1}
\end{wraptable}

We show the impact of the adversarial loss on S-WUTON. We train a variant student model S-WUTON without the adversarial loss. We provide a comparison of synthesized images in Fig. \ref{fig:vis_distillation}, and IS and FID scores in Table \ref{wrap-tab:1}. The adversarial loss on the student model is a constraint to make the student model closer to the real data distribution and to not only follow the teacher's distribution. Without the adversarial loss, the student model does not preserve person's attributes, even though they are not masked. 

\addtolength{\tabcolsep}{-6pt}
\addtolength{\tabcolsep}{6pt}

\section{Conclusion}
In this work, we propose a teacher-student setting to distill the standard virtual try-on pipeline and refocus on the initial task: changing the cloth of a non-masked person. This leads to a significant computational speed-up and largely improves image quality. Importantly, this allows to preserve person's attributes such as hands or accessories, which is necessary for a virtual try-on.

{\small \section*{Acknowledgements.} The authors thank David Picard and Marie-Morgane Paumard for their helpful advices.}

{\small
\newpage
\bibliographystyle{ieee_fullname.bst}
\bibliography{main_eccv}

\begin{thebibliography}{10}\itemsep=-1pt

\bibitem{alp2018densepose}
R{\i}za Alp~G{\"u}ler, Natalia Neverova, and Iasonas Kokkinos.
\newblock Densepose: Dense human pose estimation in the wild.
\newblock In {\em Proceedings of the IEEE Conference on Computer Vision and
  Pattern Recognition}, pages 7297--7306, 2018.

\bibitem{arjovsky2017wasserstein}
Martin Arjovsky, Soumith Chintala, and L{\'e}on Bottou.
\newblock Wasserstein generative adversarial networks.
\newblock In {\em International Conference on Machine Learning}, pages
  214--223, 2017.

\bibitem{belongie2002shape}
Serge Belongie, Jitendra Malik, and Jan Puzicha.
\newblock Shape matching and object recognition using shape contexts.
\newblock {\em IEEE Transactions on Pattern Analysis and Machine Intelligence},
  24(4):509--522, 2002.

\bibitem{bookstein1989principal}
Fred~L. Bookstein.
\newblock Principal warps: Thin-plate splines and the decomposition of
  deformations.
\newblock {\em IEEE Transactions on pattern analysis and machine intelligence},
  11(6):567--585, 1989.

\bibitem{dong2018soft}
Haoye Dong, Xiaodan Liang, Ke Gong, Hanjiang Lai, Jia Zhu, and Jian Yin.
\newblock Soft-gated warping-gan for pose-guided person image synthesis.
\newblock In {\em Advances in Neural Information Processing Systems}, pages
  472--482, 2018.

\bibitem{dong2019towards}
Haoye Dong, Xiaodan Liang, Bochao Wang, Hanjiang Lai, Jia Zhu, and Jian Yin.
\newblock Towards multi-pose guided virtual try-on network.
\newblock In {\em The IEEE International Conference on Computer Vision (ICCV)},
  October 2019.

\bibitem{gong2017look}
Ke Gong, Xiaodan Liang, Dongyu Zhang, Xiaohui Shen, and Liang Lin.
\newblock Look into person: Self-supervised structure-sensitive learning and a
  new benchmark for human parsing.
\newblock In {\em Proceedings of the IEEE Conference on Computer Vision and
  Pattern Recognition}, pages 932--940, 2017.

\bibitem{goodfellow2014generative}
Ian Goodfellow, Jean Pouget-Abadie, Mehdi Mirza, Bing Xu, David Warde-Farley,
  Sherjil Ozair, Aaron Courville, and Yoshua Bengio.
\newblock Generative adversarial nets.
\newblock In {\em Advances in neural information processing systems}, pages
  2672--2680, 2014.

\bibitem{guan2012drape}
Peng Guan, Loretta Reiss, David~A Hirshberg, Alexander Weiss, and Michael~J
  Black.
\newblock Drape: Dressing any person.
\newblock {\em ACM Trans. Graph.}, 31(4):35--1, 2012.

\bibitem{gulrajani2017improved}
Ishaan Gulrajani, Faruk Ahmed, Martin Arjovsky, Vincent Dumoulin, and Aaron~C
  Courville.
\newblock Improved training of wasserstein gans.
\newblock In {\em Advances in Neural Information Processing Systems}, pages
  5767--5777, 2017.

\bibitem{hahn2014subspace}
Fabian Hahn, Bernhard Thomaszewski, Stelian Coros, Robert~W Sumner, Forrester
  Cole, Mark Meyer, Tony DeRose, and Markus Gross.
\newblock Subspace clothing simulation using adaptive bases.
\newblock {\em ACM Transactions on Graphics (TOG)}, 33(4):105, 2014.

\bibitem{Han_2019_ICCV}
Xintong Han, Xiaojun Hu, Weilin Huang, and Matthew~R. Scott.
\newblock Clothflow: A flow-based model for clothed person generation.
\newblock In {\em The IEEE International Conference on Computer Vision (ICCV)},
  October 2019.

\bibitem{han2018viton}
Xintong Han, Zuxuan Wu, Zhe Wu, Ruichi Yu, and Larry~S Davis.
\newblock Viton: An image-based virtual try-on network.
\newblock In {\em Proceedings of the IEEE Conference on Computer Vision and
  Pattern Recognition}, pages 7543--7552, 2018.

\bibitem{heusel2017gans}
Martin Heusel, Hubert Ramsauer, Thomas Unterthiner, Bernhard Nessler, and Sepp
  Hochreiter.
\newblock Gans trained by a two time-scale update rule converge to a local nash
  equilibrium.
\newblock In {\em Advances in Neural Information Processing Systems}, pages
  6626--6637, 2017.

\bibitem{hinton2015distilling}
Geoffrey Hinton, Oriol Vinyals, and Jeff Dean.
\newblock Distilling the knowledge in a neural network.
\newblock {\em arXiv preprint arXiv:1503.02531}, 2015.

\bibitem{ioffe2015batch}
Sergey Ioffe and Christian Szegedy.
\newblock Batch normalization: Accelerating deep network training by reducing
  internal covariate shift.
\newblock In {\em International Conference on Machine Learning}, pages
  448--456, 2015.

\bibitem{isola2017image}
Phillip Isola, Jun-Yan Zhu, Tinghui Zhou, and Alexei~A Efros.
\newblock Image-to-image translation with conditional adversarial networks.
\newblock In {\em Proceedings of the IEEE conference on computer vision and
  pattern recognition}, pages 1125--1134, 2017.

\bibitem{jaderberg2015spatial}
Max Jaderberg, Karen Simonyan, Andrew Zisserman, et~al.
\newblock Spatial transformer networks.
\newblock In {\em Advances in neural information processing systems}, pages
  2017--2025, 2015.

\bibitem{jetchev2017conditional}
Nikolay Jetchev and Urs Bergmann.
\newblock The conditional analogy gan: Swapping fashion articles on people
  images.
\newblock In {\em Proceedings of the IEEE International Conference on Computer
  Vision}, pages 2287--2292, 2017.

\bibitem{johnson2016perceptual}
Justin Johnson, Alexandre Alahi, and Li Fei-Fei.
\newblock Perceptual losses for real-time style transfer and super-resolution.
\newblock In {\em European conference on computer vision}, pages 694--711.
  Springer, 2016.

\bibitem{jolicoeur-martineau2018}
Alexia Jolicoeur-Martineau.
\newblock The relativistic discriminator: a key element missing from standard
  {GAN}.
\newblock In {\em International Conference on Learning Representations}, 2019.

\bibitem{kingma2014adam}
Diederik~P Kingma and Jimmy Ba.
\newblock Adam: A method for stochastic optimization.
\newblock In {\em International Conference on Learning Representations}, 2015.

\bibitem{liang2019look}
Xiaodan Liang, Ke Gong, Xiaohui Shen, and Liang Lin.
\newblock Look into person: Joint body parsing \& pose estimation network and a
  new benchmark.
\newblock {\em IEEE transactions on pattern analysis and machine intelligence},
  41(4):871--885, 2019.

\bibitem{lorenz2019unsupervised}
Dominik Lorenz, Leonard Bereska, Timo Milbich, and Bj{\"o}rn Ommer.
\newblock Unsupervised part-based disentangling of object shape and appearance.
\newblock In {\em Proceedings of the IEEE Conference on Computer Vision and
  Pattern Recognition}, 2019.

\bibitem{ma2018disentangled}
Liqian Ma, Qianru Sun, Stamatios Georgoulis, Luc Van~Gool, Bernt Schiele, and
  Mario Fritz.
\newblock Disentangled person image generation.
\newblock In {\em Proceedings of the IEEE Conference on Computer Vision and
  Pattern Recognition}, pages 99--108, 2018.

\bibitem{mejjati2018unsupervised}
Youssef~Alami Mejjati, Christian Richardt, James Tompkin, Darren Cosker, and
  Kwang~In Kim.
\newblock Unsupervised attention-guided image-to-image translation.
\newblock In {\em Advances in Neural Information Processing Systems}, pages
  3693--3703, 2018.

\bibitem{mirza2014conditional}
Mehdi Mirza and Simon Osindero.
\newblock Conditional generative adversarial nets.
\newblock {\em arXiv preprint arXiv:1411.1784}, 2014.

\bibitem{pons2017clothcap}
Gerard Pons-Moll, Sergi Pujades, Sonny Hu, and Michael~J Black.
\newblock Clothcap: Seamless 4d clothing capture and retargeting.
\newblock {\em ACM Transactions on Graphics (TOG)}, 36(4):73, 2017.

\bibitem{radford2015unsupervised}
Alec Radford, Luke Metz, and Soumith Chintala.
\newblock Unsupervised representation learning with deep convolutional
  generative adversarial networks.
\newblock In {\em International Conference on Learning Representations}, 2016.

\bibitem{raj2018swapnet}
Amit Raj, Patsorn Sangkloy, Huiwen Chang, Jingwan Lu, Duygu Ceylan, and James
  Hays.
\newblock Swapnet: Garment transfer in single view images.
\newblock In {\em Proceedings of the European Conference on Computer Vision
  (ECCV)}, pages 666--682, 2018.

\bibitem{rocco2017convolutional}
Ignacio Rocco, Relja Arandjelovic, and Josef Sivic.
\newblock Convolutional neural network architecture for geometric matching.
\newblock In {\em Proceedings of the IEEE Conference on Computer Vision and
  Pattern Recognition}, pages 6148--6157, 2017.

\bibitem{ronneberger2015u}
Olaf Ronneberger, Philipp Fischer, and Thomas Brox.
\newblock U-net: Convolutional networks for biomedical image segmentation.
\newblock In {\em International Conference on Medical image computing and
  computer-assisted intervention}, pages 234--241. Springer, 2015.

\bibitem{salimans2016improved}
Tim Salimans, Ian Goodfellow, Wojciech Zaremba, Vicki Cheung, Alec Radford, and
  Xi Chen.
\newblock Improved techniques for training gans.
\newblock In {\em Advances in neural information processing systems}, pages
  2234--2242, 2016.

\bibitem{shetty2018adversarial}
Rakshith~R Shetty, Mario Fritz, and Bernt Schiele.
\newblock Adversarial scene editing: Automatic object removal from weak
  supervision.
\newblock In {\em Advances in Neural Information Processing Systems}, pages
  7717--7727, 2018.

\bibitem{siarohin2018deformable}
Aliaksandr Siarohin, Enver Sangineto, St{\'e}phane Lathuili{\`e}re, and Nicu
  Sebe.
\newblock Deformable gans for pose-based human image generation.
\newblock In {\em Proceedings of the IEEE Conference on Computer Vision and
  Pattern Recognition}, pages 3408--3416, 2018.

\bibitem{simonyan2014very}
Karen Simonyan and Andrew Zisserman.
\newblock Very deep convolutional networks for large-scale image recognition.
\newblock In {\em International Conference on Learning Representations}, 2015.

\bibitem{ulyanov2017improved}
Dmitry Ulyanov, Andrea Vedaldi, and Victor Lempitsky.
\newblock Improved texture networks: Maximizing quality and diversity in
  feed-forward stylization and texture synthesis.
\newblock In {\em Proceedings of the IEEE Conference on Computer Vision and
  Pattern Recognition}, pages 6924--6932, 2017.

\bibitem{wang2018toward}
Bochao Wang, Huabin Zheng, Xiaodan Liang, Yimin Chen, Liang Lin, and Meng Yang.
\newblock Toward characteristic-preserving image-based virtual try-on network.
\newblock In {\em Proceedings of the European Conference on Computer Vision
  (ECCV)}, pages 589--604, 2018.

\bibitem{wang2004image}
Zhou Wang, Alan~C Bovik, Hamid~R Sheikh, and Eero~P Simoncelli.
\newblock Image quality assessment: from error visibility to structural
  similarity.
\newblock {\em IEEE transactions on image processing}, 13(4):600--612, 2004.

\bibitem{wu2018m2e}
Zhonghua Wu, Guosheng Lin, Qingyi Tao, and Jianfei Cai.
\newblock M2e-try on net: Fashion from model to everyone.
\newblock {\em arXiv preprint arXiv:1811.08599}, 2018.

\bibitem{yang2020towards}
Han Yang, Ruimao Zhang, Xiaobao Guo, Wei Liu, Wangmeng Zuo, and Ping Luo.
\newblock Towards photo-realistic virtual try-on by adaptively
  generating-preserving image content.
\newblock In {\em Proceedings of the IEEE/CVF Conference on Computer Vision and
  Pattern Recognition}, pages 7850--7859, 2020.

\bibitem{Yu_2019_ICCV}
Ruiyun Yu, Xiaoqi Wang, and Xiaohui Xie.
\newblock Vtnfp: An image-based virtual try-on network with body and clothing
  feature preservation.
\newblock In {\em The IEEE International Conference on Computer Vision (ICCV)},
  October 2019.

\bibitem{zhang2018unreasonable}
Richard Zhang, Phillip Isola, Alexei~A Efros, Eli Shechtman, and Oliver Wang.
\newblock The unreasonable effectiveness of deep features as a perceptual
  metric.
\newblock In {\em Proceedings of the IEEE Conference on Computer Vision and
  Pattern Recognition}, pages 586--595, 2018.

\bibitem{zhao2016loss}
Hang Zhao, Orazio Gallo, Iuri Frosio, and Jan Kautz.
\newblock Loss functions for image restoration with neural networks.
\newblock {\em IEEE Transactions on computational imaging}, 3(1):47--57, 2016.

\bibitem{zhu2017unpaired}
Jun-Yan Zhu, Taesung Park, Phillip Isola, and Alexei~A Efros.
\newblock Unpaired image-to-image translation using cycle-consistent
  adversarial networks.
\newblock In {\em Proceedings of the IEEE International Conference on Computer
  Vision}, pages 2223--2232, 2017.

\end{thebibliography}
}

\clearpage
\newpage


\section{Appendix}
\subsection{Implementation details}
\textbf{Convolutional geometric matcher.} To extract the feature maps, we apply five times one standard convolution layer followed by a 2-strided convolution layer which downsamples the maps. The depth of the feature maps at each scale is (16,32,64,128,256). The correlation map is then computed and feeds a regression network composed of two 2-strided convolution layers, two standard convolution layers and one final fully connected layer predicting a vector $\theta \in \mathbb{R}^{50}$. We use batch normalization \cite{ioffe2015batch} and relu activation. The parameters of the two feature maps extractors are not shared.

\textbf{Siamese U-net generator.} We use the same encoder architecture as in the convolutional geometric matcher, but we store the feature maps at each scale. The decoder has an architecture symmetric to the encoder. There are five times one standard convolution layer followed by a 2-strided deconvolution layer which upsamples the feature maps. After a deconvolution, the feature maps are concatenated with the feature maps passed through the skip connections. In the generator, we use instance normalization, which shows better results for image and texture generation \cite{ulyanov2017improved}, with relu activation.

\textbf{Discriminator.} We adopt the fully convolutional discriminator from Pix-2-Pix \cite{isola2017image}, but with five downsampling layers instead of three in the original version. Each of it is composed of: 2-strided convolution, batch normalization, leaky relu, 1-strided convolution, batch normalization, leaky relu.

\textbf{Adversarial loss.} We use the relativistic formulation of the adversarial loss \cite{jolicoeur-martineau2018}. In this formulation, the discriminator is trained to predict that real images are more real than synthesized ones, rather than trained to predict that real images are real and synthesized images are synthesized. We also use gradient penalty on the discriminator.

\textbf{Optimization.} We use the Adam optimizer \cite{kingma2014adam} with $\beta_1 = 0.5$, $\beta_2 = 0.999$, a learning rate of $10e^{-3}$ and a batch size of 8. Also, we use $\lambda_p = \lambda_{L1} = \lambda_w = \lambda_{adv} = 1$.

\textbf{Hardware.} We use a NVIDIA Tesla V100 with 16GB of RAM. The training takes around 2 days for T-WUTON, and around 3 days for S-WUTON. For inference, S-WUTON processes $\sim$77 frames per second.

\subsection{More results from S-WUTON}
\begin{figure}[]
\fontsize{8}{8}\selectfont
\centering
\begin{tabular*}{340pt}{@{\extracolsep{\fill}}cccccccccccc}
Cloth/Person &
\subfigure{\includegraphics[width=0.080\linewidth]{demo_S_WUTON/4BE21D09U-Q11_10=person_half_front.jpg}} &
\subfigure{\includegraphics[width=0.080\linewidth]{demo_S_WUTON/AN621DA8W-M11_10=person_half_front.jpg}} &
\subfigure{\includegraphics[width=0.080\linewidth]{demo_S_WUTON/AN621DA75-J11_10=person_half_front.jpg}} &
\subfigure{\includegraphics[width=0.080\linewidth]{demo_S_WUTON/AN621DA82-C11_8=person_half_front.jpg}} &
\subfigure{\includegraphics[width=0.080\linewidth]{demo_S_WUTON/BX329G012-Q11_11=person_whole_front.jpg}} &
\subfigure{\includegraphics[width=0.080\linewidth]{demo_S_WUTON/CH621D03C-A11_10=person_half_front.jpg}} &
\subfigure{\includegraphics[width=0.080\linewidth]{demo_S_WUTON/DP521E0QM-J11_10=person_half_front.jpg}} &
\subfigure{\includegraphics[width=0.080\linewidth]{demo_S_WUTON/ED121D0OT-K11_9=person_whole_front.jpg}} &
\subfigure{\includegraphics[width=0.080\linewidth]{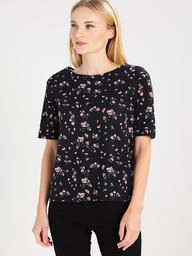}} &
\subfigure{\includegraphics[width=0.080\linewidth]{demo_S_WUTON/ES121D0MX-K11_10=person_half_front.jpg}}\\ [-2.2ex]
\subfigure{\includegraphics[width=0.080\linewidth]{demo_S_WUTON/0VB21E007-T11_10=cloth_front.jpg}} &
\subfigure{\includegraphics[width=0.080\linewidth]{demo_S_WUTON/4BE21D09U-Q11_10=person_half_front+0VB21E007-T11_10=cloth_front.jpg}} &
\subfigure{\includegraphics[width=0.080\linewidth]{demo_S_WUTON/AN621DA8W-M11_10=person_half_front+0VB21E007-T11_10=cloth_front.jpg}} &
\subfigure{\includegraphics[width=0.080\linewidth]{demo_S_WUTON/AN621DA75-J11_10=person_half_front+0VB21E007-T11_10=cloth_front.jpg}} &
\subfigure{\includegraphics[width=0.080\linewidth]{demo_S_WUTON/AN621DA82-C11_8=person_half_front+0VB21E007-T11_10=cloth_front.jpg}} &
\subfigure{\includegraphics[width=0.080\linewidth]{demo_S_WUTON/BX329G012-Q11_11=person_whole_front+0VB21E007-T11_10=cloth_front.jpg}} &
\subfigure{\includegraphics[width=0.080\linewidth]{demo_S_WUTON/CH621D03C-A11_10=person_half_front+0VB21E007-T11_10=cloth_front.jpg}} &
\subfigure{\includegraphics[width=0.080\linewidth]{demo_S_WUTON/DP521E0QM-J11_10=person_half_front+0VB21E007-T11_10=cloth_front.jpg}} &
\subfigure{\includegraphics[width=0.080\linewidth]{demo_S_WUTON/ED121D0OT-K11_9=person_whole_front+0VB21E007-T11_10=cloth_front.jpg}} &
\subfigure{\includegraphics[width=0.080\linewidth]{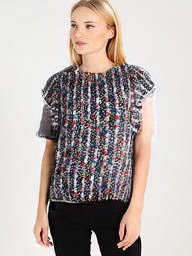}} &
\subfigure{\includegraphics[width=0.080\linewidth]{demo_S_WUTON/ES121D0MX-K11_10=person_half_front+0VB21E007-T11_10=cloth_front.jpg}} \\ [-2.2ex]
\subfigure{\includegraphics[width=0.080\linewidth]{demo_S_WUTON/1VJ21E00C-K11_10=cloth_front.jpg}} &
\subfigure{\includegraphics[width=0.080\linewidth]{demo_S_WUTON/4BE21D09U-Q11_10=person_half_front+1VJ21E00C-K11_10=cloth_front.jpg}} &
\subfigure{\includegraphics[width=0.080\linewidth]{demo_S_WUTON/AN621DA8W-M11_10=person_half_front+1VJ21E00C-K11_10=cloth_front.jpg}} &
\subfigure{\includegraphics[width=0.080\linewidth]{demo_S_WUTON/AN621DA75-J11_10=person_half_front+1VJ21E00C-K11_10=cloth_front.jpg}} &
\subfigure{\includegraphics[width=0.080\linewidth]{demo_S_WUTON/AN621DA82-C11_8=person_half_front+1VJ21E00C-K11_10=cloth_front.jpg}} &
\subfigure{\includegraphics[width=0.080\linewidth]{demo_S_WUTON/BX329G012-Q11_11=person_whole_front+1VJ21E00C-K11_10=cloth_front.jpg}} &
\subfigure{\includegraphics[width=0.080\linewidth]{demo_S_WUTON/CH621D03C-A11_10=person_half_front+1VJ21E00C-K11_10=cloth_front.jpg}} &
\subfigure{\includegraphics[width=0.080\linewidth]{demo_S_WUTON/DP521E0QM-J11_10=person_half_front+1VJ21E00C-K11_10=cloth_front.jpg}} &
\subfigure{\includegraphics[width=0.080\linewidth]{demo_S_WUTON/ED121D0OT-K11_9=person_whole_front+1VJ21E00C-K11_10=cloth_front.jpg}} &
\subfigure{\includegraphics[width=0.080\linewidth]{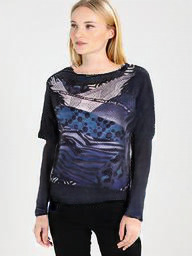}} &
\subfigure{\includegraphics[width=0.080\linewidth]{demo_S_WUTON/ES121D0MX-K11_10=person_half_front+1VJ21E00C-K11_10=cloth_front.jpg}} \\ [-2.2ex]
\subfigure{\includegraphics[width=0.080\linewidth]{demo_S_WUTON/A0F21E00V-G11_10=cloth_front.jpg}} &
\subfigure{\includegraphics[width=0.080\linewidth]{demo_S_WUTON/4BE21D09U-Q11_10=person_half_front+A0F21E00V-G11_10=cloth_front.jpg}} &
\subfigure{\includegraphics[width=0.080\linewidth]{demo_S_WUTON/AN621DA8W-M11_10=person_half_front+A0F21E00V-G11_10=cloth_front.jpg}} &
\subfigure{\includegraphics[width=0.080\linewidth]{demo_S_WUTON/AN621DA75-J11_10=person_half_front+A0F21E00V-G11_10=cloth_front.jpg}} &
\subfigure{\includegraphics[width=0.080\linewidth]{demo_S_WUTON/AN621DA82-C11_8=person_half_front+A0F21E00V-G11_10=cloth_front.jpg}} &
\subfigure{\includegraphics[width=0.080\linewidth]{demo_S_WUTON/BX329G012-Q11_11=person_whole_front+A0F21E00V-G11_10=cloth_front.jpg}} &
\subfigure{\includegraphics[width=0.080\linewidth]{demo_S_WUTON/CH621D03C-A11_10=person_half_front+A0F21E00V-G11_10=cloth_front.jpg}} &
\subfigure{\includegraphics[width=0.080\linewidth]{demo_S_WUTON/DP521E0QM-J11_10=person_half_front+A0F21E00V-G11_10=cloth_front.jpg}} &
\subfigure{\includegraphics[width=0.080\linewidth]{demo_S_WUTON/ED121D0OT-K11_9=person_whole_front+A0F21E00V-G11_10=cloth_front.jpg}} &
\subfigure{\includegraphics[width=0.080\linewidth]{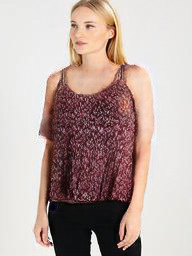}} &
\subfigure{\includegraphics[width=0.080\linewidth]{demo_S_WUTON/ES121D0MX-K11_10=person_half_front+A0F21E00V-G11_10=cloth_front.jpg}} \\ [-2.2ex]
\subfigure{\includegraphics[width=0.080\linewidth]{demo_S_WUTON/AD121D0H0-T11_10=cloth_front.jpg}} &
\subfigure{\includegraphics[width=0.080\linewidth]{demo_S_WUTON/4BE21D09U-Q11_10=person_half_front+AD121D0H0-T11_10=cloth_front.jpg}} &
\subfigure{\includegraphics[width=0.080\linewidth]{demo_S_WUTON/AN621DA8W-M11_10=person_half_front+AD121D0H0-T11_10=cloth_front.jpg}} &
\subfigure{\includegraphics[width=0.080\linewidth]{demo_S_WUTON/AN621DA75-J11_10=person_half_front+AD121D0H0-T11_10=cloth_front.jpg}} &
\subfigure{\includegraphics[width=0.080\linewidth]{demo_S_WUTON/AN621DA82-C11_8=person_half_front+AD121D0H0-T11_10=cloth_front.jpg}} &
\subfigure{\includegraphics[width=0.080\linewidth]{demo_S_WUTON/BX329G012-Q11_11=person_whole_front+AD121D0H0-T11_10=cloth_front.jpg}} &
\subfigure{\includegraphics[width=0.080\linewidth]{demo_S_WUTON/CH621D03C-A11_10=person_half_front+AD121D0H0-T11_10=cloth_front.jpg}} &
\subfigure{\includegraphics[width=0.080\linewidth]{demo_S_WUTON/DP521E0QM-J11_10=person_half_front+AD121D0H0-T11_10=cloth_front.jpg}} &
\subfigure{\includegraphics[width=0.080\linewidth]{demo_S_WUTON/ED121D0OT-K11_9=person_whole_front+AD121D0H0-T11_10=cloth_front.jpg}} &
\subfigure{\includegraphics[width=0.080\linewidth]{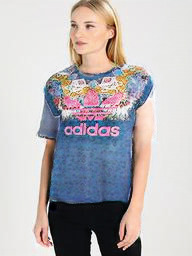}} &
\subfigure{\includegraphics[width=0.080\linewidth]{demo_S_WUTON/ES121D0MX-K11_10=person_half_front+AD121D0H0-T11_10=cloth_front.jpg}} \\ [-2.2ex]
\subfigure{\includegraphics[width=0.080\linewidth]{demo_S_WUTON/ED121D0OK-K11_12=cloth_front.jpg}} &
\subfigure{\includegraphics[width=0.080\linewidth]{demo_S_WUTON/4BE21D09U-Q11_10=person_half_front+ED121D0OK-K11_12=cloth_front.jpg}} &
\subfigure{\includegraphics[width=0.080\linewidth]{demo_S_WUTON/AN621DA8W-M11_10=person_half_front+ED121D0OK-K11_12=cloth_front.jpg}} &
\subfigure{\includegraphics[width=0.080\linewidth]{demo_S_WUTON/AN621DA75-J11_10=person_half_front+ED121D0OK-K11_12=cloth_front.jpg}} &
\subfigure{\includegraphics[width=0.080\linewidth]{demo_S_WUTON/AN621DA82-C11_8=person_half_front+ED121D0OK-K11_12=cloth_front.jpg}} &
\subfigure{\includegraphics[width=0.080\linewidth]{demo_S_WUTON/BX329G012-Q11_11=person_whole_front+ED121D0OK-K11_12=cloth_front.jpg}} &
\subfigure{\includegraphics[width=0.080\linewidth]{demo_S_WUTON/CH621D03C-A11_10=person_half_front+ED121D0OK-K11_12=cloth_front.jpg}} &
\subfigure{\includegraphics[width=0.080\linewidth]{demo_S_WUTON/DP521E0QM-J11_10=person_half_front+ED121D0OK-K11_12=cloth_front.jpg}} &
\subfigure{\includegraphics[width=0.080\linewidth]{demo_S_WUTON/ED121D0OT-K11_9=person_whole_front+ED121D0OK-K11_12=cloth_front.jpg}} &
\subfigure{\includegraphics[width=0.080\linewidth]{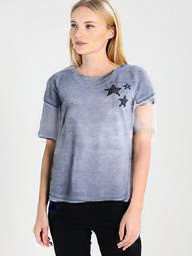}} &
\subfigure{\includegraphics[width=0.080\linewidth]{demo_S_WUTON/ES121D0MX-K11_10=person_half_front+ED121D0OK-K11_12=cloth_front.jpg}} \\ [-2.2ex]
\subfigure{\includegraphics[width=0.080\linewidth]{demo_S_WUTON/A0F21D012-K11_10=cloth_front.jpg}} &
\subfigure{\includegraphics[width=0.080\linewidth]{demo_S_WUTON/4BE21D09U-Q11_10=person_half_front+A0F21D012-K11_10=cloth_front.jpg}} &
\subfigure{\includegraphics[width=0.080\linewidth]{demo_S_WUTON/AN621DA8W-M11_10=person_half_front+A0F21D012-K11_10=cloth_front.jpg}} &
\subfigure{\includegraphics[width=0.080\linewidth]{demo_S_WUTON/AN621DA75-J11_10=person_half_front+A0F21D012-K11_10=cloth_front.jpg}} &
\subfigure{\includegraphics[width=0.080\linewidth]{demo_S_WUTON/AN621DA82-C11_8=person_half_front+A0F21D012-K11_10=cloth_front.jpg}} &
\subfigure{\includegraphics[width=0.080\linewidth]{demo_S_WUTON/BX329G012-Q11_11=person_whole_front+A0F21D012-K11_10=cloth_front.jpg}} &
\subfigure{\includegraphics[width=0.080\linewidth]{demo_S_WUTON/CH621D03C-A11_10=person_half_front+A0F21D012-K11_10=cloth_front.jpg}} &
\subfigure{\includegraphics[width=0.080\linewidth]{demo_S_WUTON/DP521E0QM-J11_10=person_half_front+A0F21D012-K11_10=cloth_front.jpg}} &
\subfigure{\includegraphics[width=0.080\linewidth]{demo_S_WUTON/ED121D0OT-K11_9=person_whole_front+A0F21D012-K11_10=cloth_front.jpg}} &
\subfigure{\includegraphics[width=0.080\linewidth]{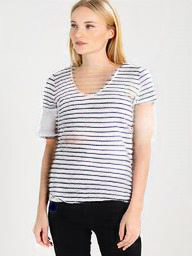}} &
\subfigure{\includegraphics[width=0.080\linewidth]{demo_S_WUTON/ES121D0MX-K11_10=person_half_front+A0F21D012-K11_10=cloth_front.jpg}} \\ [-2.2ex]
\subfigure{\includegraphics[width=0.080\linewidth]{demo_S_WUTON/ES121D0N1-K11_10=cloth_front.jpg}} &
\subfigure{\includegraphics[width=0.080\linewidth]{demo_S_WUTON/4BE21D09U-Q11_10=person_half_front+ES121D0N1-K11_10=cloth_front.jpg}} &
\subfigure{\includegraphics[width=0.080\linewidth]{demo_S_WUTON/AN621DA8W-M11_10=person_half_front+ES121D0N1-K11_10=cloth_front.jpg}} &
\subfigure{\includegraphics[width=0.080\linewidth]{demo_S_WUTON/AN621DA75-J11_10=person_half_front+ES121D0N1-K11_10=cloth_front.jpg}} &
\subfigure{\includegraphics[width=0.080\linewidth]{demo_S_WUTON/AN621DA82-C11_8=person_half_front+ES121D0N1-K11_10=cloth_front.jpg}} &
\subfigure{\includegraphics[width=0.080\linewidth]{demo_S_WUTON/BX329G012-Q11_11=person_whole_front+ES121D0N1-K11_10=cloth_front.jpg}} &
\subfigure{\includegraphics[width=0.080\linewidth]{demo_S_WUTON/CH621D03C-A11_10=person_half_front+ES121D0N1-K11_10=cloth_front.jpg}} &
\subfigure{\includegraphics[width=0.080\linewidth]{demo_S_WUTON/DP521E0QM-J11_10=person_half_front+ES121D0N1-K11_10=cloth_front.jpg}} &
\subfigure{\includegraphics[width=0.080\linewidth]{demo_S_WUTON/ED121D0OT-K11_9=person_whole_front+ES121D0N1-K11_10=cloth_front.jpg}} &
\subfigure{\includegraphics[width=0.080\linewidth]{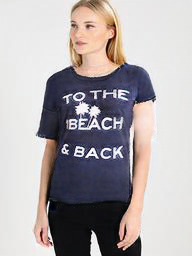}} &
\subfigure{\includegraphics[width=0.080\linewidth]{demo_S_WUTON/ES121D0MX-K11_10=person_half_front+ES121D0N1-K11_10=cloth_front.jpg}} \\ [-2.2ex]
\subfigure{\includegraphics[width=0.080\linewidth]{demo_S_WUTON/ES421D09P-Q11_12=cloth_front.jpg}} &
\subfigure{\includegraphics[width=0.080\linewidth]{demo_S_WUTON/4BE21D09U-Q11_10=person_half_front+ES421D09P-Q11_12=cloth_front.jpg}} &
\subfigure{\includegraphics[width=0.080\linewidth]{demo_S_WUTON/AN621DA8W-M11_10=person_half_front+ES421D09P-Q11_12=cloth_front.jpg}} &
\subfigure{\includegraphics[width=0.080\linewidth]{demo_S_WUTON/AN621DA75-J11_10=person_half_front+ES421D09P-Q11_12=cloth_front.jpg}} &
\subfigure{\includegraphics[width=0.080\linewidth]{demo_S_WUTON/AN621DA82-C11_8=person_half_front+ES421D09P-Q11_12=cloth_front.jpg}} &
\subfigure{\includegraphics[width=0.080\linewidth]{demo_S_WUTON/BX329G012-Q11_11=person_whole_front+ES421D09P-Q11_12=cloth_front.jpg}} &
\subfigure{\includegraphics[width=0.080\linewidth]{demo_S_WUTON/CH621D03C-A11_10=person_half_front+ES421D09P-Q11_12=cloth_front.jpg}} &
\subfigure{\includegraphics[width=0.080\linewidth]{demo_S_WUTON/DP521E0QM-J11_10=person_half_front+ES421D09P-Q11_12=cloth_front.jpg}} &
\subfigure{\includegraphics[width=0.080\linewidth]{demo_S_WUTON/ED121D0OT-K11_9=person_whole_front+ES421D09P-Q11_12=cloth_front.jpg}} &
\subfigure{\includegraphics[width=0.080\linewidth]{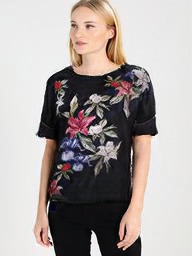}} &
\subfigure{\includegraphics[width=0.080\linewidth]{demo_S_WUTON/ES121D0MX-K11_10=person_half_front+ES421D09P-Q11_12=cloth_front.jpg}} \\ [-2.2ex]
\subfigure{\includegraphics[width=0.080\linewidth]{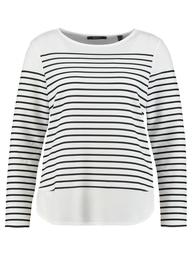}} &
\subfigure{\includegraphics[width=0.080\linewidth]{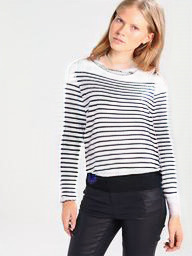}} &
\subfigure{\includegraphics[width=0.080\linewidth]{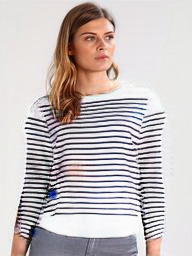}} &
\subfigure{\includegraphics[width=0.080\linewidth]{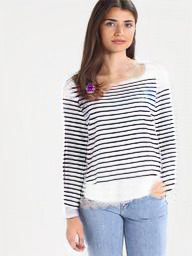}} &
\subfigure{\includegraphics[width=0.080\linewidth]{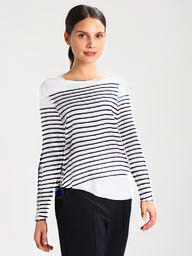}} &
\subfigure{\includegraphics[width=0.080\linewidth]{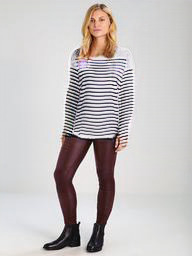}} &
\subfigure{\includegraphics[width=0.080\linewidth]{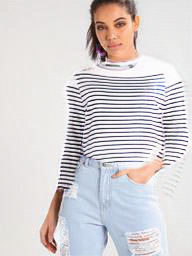}} &
\subfigure{\includegraphics[width=0.080\linewidth]{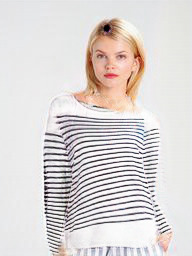}} &
\subfigure{\includegraphics[width=0.080\linewidth]{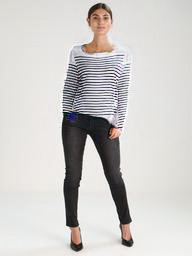}} &
\subfigure{\includegraphics[width=0.080\linewidth]{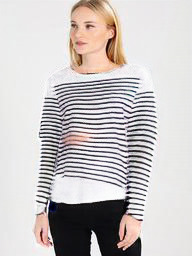}} &
\subfigure{\includegraphics[width=0.080\linewidth]{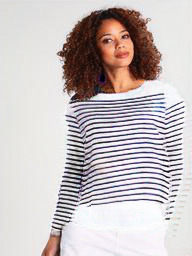}} \\ [-2.2ex]
\subfigure{\includegraphics[width=0.080\linewidth]{demo_S_WUTON/ED121D0HY-A11_12=cloth_front.jpg}} &
\subfigure{\includegraphics[width=0.080\linewidth]{demo_S_WUTON/4BE21D09U-Q11_10=person_half_front+ED121D0HY-A11_12=cloth_front.jpg}} &
\subfigure{\includegraphics[width=0.080\linewidth]{demo_S_WUTON/AN621DA8W-M11_10=person_half_front+ED121D0HY-A11_12=cloth_front.jpg}} &
\subfigure{\includegraphics[width=0.080\linewidth]{demo_S_WUTON/AN621DA75-J11_10=person_half_front+ED121D0HY-A11_12=cloth_front.jpg}} &
\subfigure{\includegraphics[width=0.080\linewidth]{demo_S_WUTON/AN621DA82-C11_8=person_half_front+ED121D0HY-A11_12=cloth_front.jpg}} &
\subfigure{\includegraphics[width=0.080\linewidth]{demo_S_WUTON/BX329G012-Q11_11=person_whole_front+ED121D0HY-A11_12=cloth_front.jpg}} &
\subfigure{\includegraphics[width=0.080\linewidth]{demo_S_WUTON/CH621D03C-A11_10=person_half_front+ED121D0HY-A11_12=cloth_front.jpg}} &
\subfigure{\includegraphics[width=0.080\linewidth]{demo_S_WUTON/DP521E0QM-J11_10=person_half_front+ED121D0HY-A11_12=cloth_front.jpg}} &
\subfigure{\includegraphics[width=0.080\linewidth]{demo_S_WUTON/ED121D0OT-K11_9=person_whole_front+ED121D0HY-A11_12=cloth_front.jpg}} &
\subfigure{\includegraphics[width=0.080\linewidth]{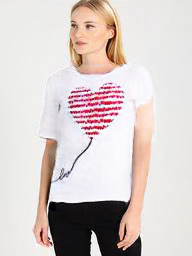}} &
\subfigure{\includegraphics[width=0.080\linewidth]{demo_S_WUTON/ES121D0MX-K11_10=person_half_front+ED121D0HY-A11_12=cloth_front.jpg}} \\ [-2.2ex]
\end{tabular*}
\caption{\label{fig:appendix_cross_examples} More visual results from S-WUTON.}
\end{figure}

We show more results from S-WUTON model in Fig. \ref{fig:appendix_cross_examples}. It shows the abilities of our model to preserve complex cloth patterns (stripes, text or textures) and body details. It is robust across a wide range of human pose. On the antepenultimate column, we show a common failure case of our method, when sleeves of the source person are too large.
\addtolength{\tabcolsep}{-6pt}
\addtolength{\tabcolsep}{6pt}

\subsection{More visual examples on the importance of distillation}
\begin{figure}
\fontsize{8}{8}\selectfont
\centering
\begin{tabular*}{340pt}{@{\extracolsep{\fill}}cccccc}
Reference & Target & Human & CP-VTON & T-WUTON & S-WUTON \\
person & cloth & parsing &  & (ours) & (ours) \\ [-0.2ex]
\subfigure{\includegraphics[width=0.16\linewidth]{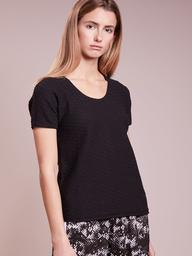}} &
\subfigure{\includegraphics[width=0.16\linewidth]{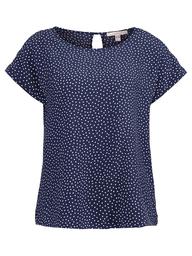}} &
\subfigure{\includegraphics[width=0.16\linewidth]{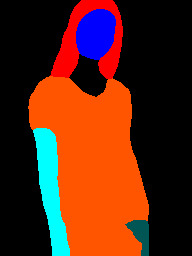}} &
\subfigure{\includegraphics[width=0.16\linewidth]{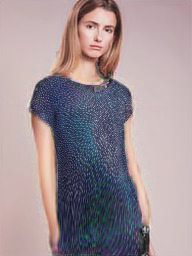}} &
\subfigure{\includegraphics[width=0.16\linewidth]{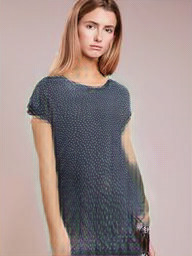}} &
\subfigure{\includegraphics[width=0.16\linewidth]{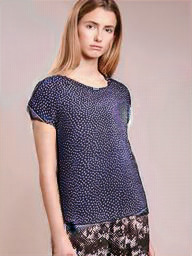}}\\ [-2.6ex]
\subfigure{\includegraphics[width=0.16\linewidth]{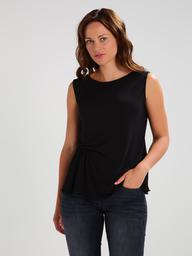}} &
\subfigure{\includegraphics[width=0.16\linewidth]{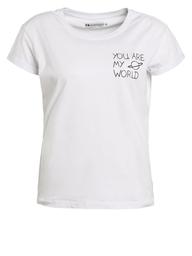}} &
\subfigure{\includegraphics[width=0.16\linewidth]{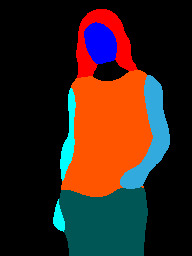}} &
\subfigure{\includegraphics[width=0.16\linewidth]{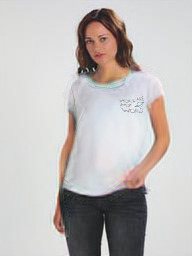}} &
\subfigure{\includegraphics[width=0.16\linewidth]{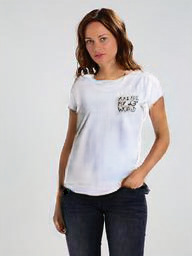}} &
\subfigure{\includegraphics[width=0.16\linewidth]{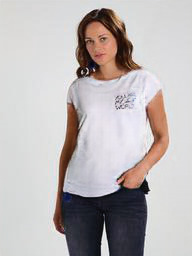}}\\ [-2.6ex]
\subfigure{\includegraphics[width=0.16\linewidth]{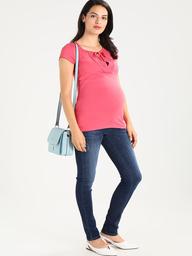}} &
\subfigure{\includegraphics[width=0.16\linewidth]{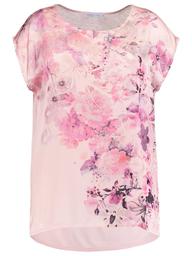}} &
\subfigure{\includegraphics[width=0.16\linewidth]{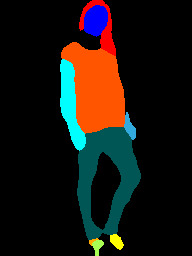}} &
\subfigure{\includegraphics[width=0.16\linewidth]{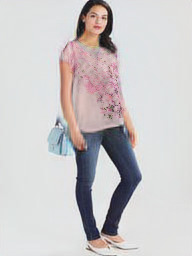}} &
\subfigure{\includegraphics[width=0.16\linewidth]{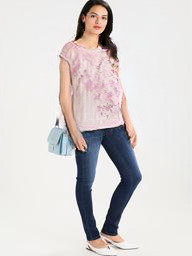}} &
\subfigure{\includegraphics[width=0.16\linewidth]{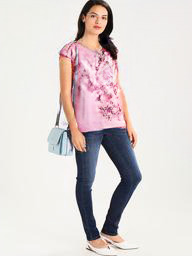}}\\ [-2.6ex]
\subfigure{\includegraphics[width=0.16\linewidth]{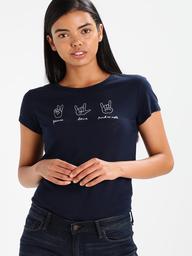}} &
\subfigure{\includegraphics[width=0.16\linewidth]{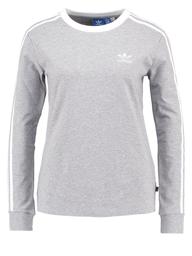}} &
\subfigure{\includegraphics[width=0.16\linewidth]{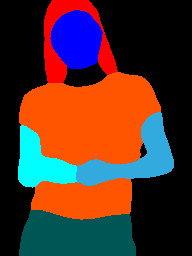}} &
\subfigure{\includegraphics[width=0.16\linewidth]{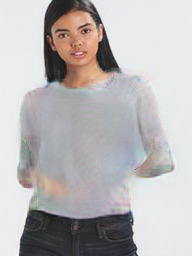}} &
\subfigure{\includegraphics[width=0.16\linewidth]{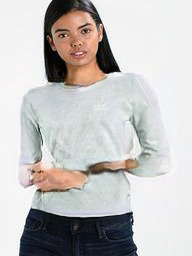}} &
\subfigure{\includegraphics[width=0.16\linewidth]{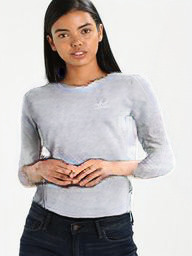}}\\  [-2.6ex]
\subfigure{\includegraphics[width=0.16\linewidth]{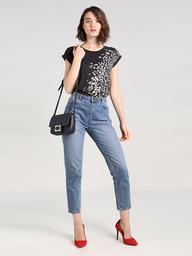}} &
\subfigure{\includegraphics[width=0.16\linewidth]{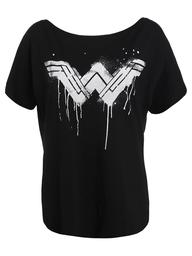}} &
\subfigure{\includegraphics[width=0.16\linewidth]{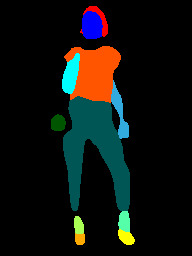}} &
\subfigure{\includegraphics[width=0.16\linewidth]{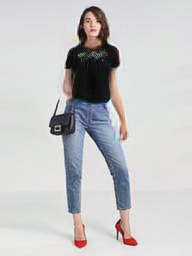}} &
\subfigure{\includegraphics[width=0.16\linewidth]{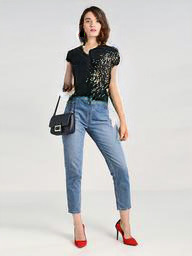}} &
\subfigure{\includegraphics[width=0.16\linewidth]{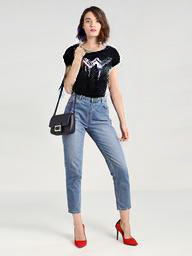}}\\
[-2.6ex]
\subfigure{\includegraphics[width=0.16\linewidth]{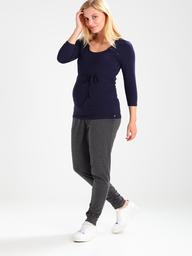}} &
\subfigure{\includegraphics[width=0.16\linewidth]{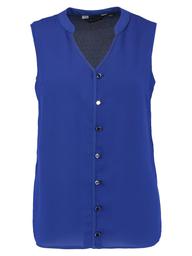}} &
\subfigure{\includegraphics[width=0.16\linewidth]{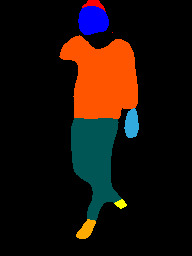}} &
\subfigure{\includegraphics[width=0.16\linewidth]{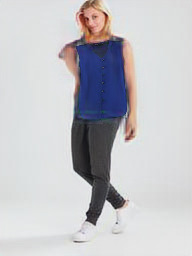}} &
\subfigure{\includegraphics[width=0.16\linewidth]{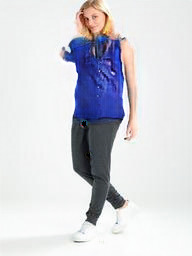}} &
\subfigure{\includegraphics[width=0.16\linewidth]{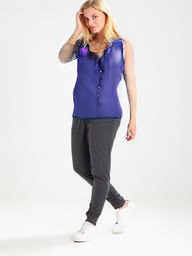}}\\[-0.6ex]
\end{tabular*}
\caption{\label{fig:appendix_distillation} Visual results proving the importance of the student-teacher approach. It is robust to parsing errors and preserves person's attributes such as arms, hands, and handbags.}
\end{figure}

In Fig. \ref{fig:appendix_distillation}, we show more visual results proving the soundness of our teacher-student approach.

Visually, our student model solves two kinds of problems: it is robust to human parser errors; it preserves important information that is masked to the standard virtual try-ons (hands, skin, handbags).
\addtolength{\tabcolsep}{-6pt}
\addtolength{\tabcolsep}{6pt}

\newpage

\subsection{Comparisons with VTNFP \cite{Yu_2019_ICCV}}
In Fig. \ref{fig:appendix_VTNFP}, we show visual comparisons between CP-VTON, VTNFP, T-WUTON and S-WUTON. Images from all columns except T-WUTON and S-WUTON are taken from their paper, which explains the low resolution. Note that they trained their model on the original VITON dataset, that is now forbidden due to copyright issues. As mentioned in the paper, our model is trained on the dataset released in MG-VTON \cite{dong2019towards}.

Results show that S-WUTON produces sharper images, with body details better preserved (especially the hands). Cloth patterns are also better rendered with S-WUTON (e.g. row 2). However, their model handles better difficult poses, when models are crossing arms (e.g rows 5,6,8). Note that our model performs well on persons crossing arms on MG-VTON dataset (see paper and Fig. \ref{fig:appendix_distillation} of Appendix).

In terms of computational cost at inference time, our S-WUTON is at least 13 times faster than their model. Moreover, since their pipeline relies on human parsing and pose estimation, it is also sensitive to the errors exhibited in our paper and in Fig. \ref{fig:appendix_distillation} of the Appendix.

\addtolength{\tabcolsep}{-6pt}
\begin{figure}[]
\fontsize{8}{8}\selectfont
\centering
\begin{tabular*}{340pt}{@{\extracolsep{\fill}}cccccc}
Reference & Target & CP-VTON & VTNFP & T-WUTON & S-WUTON\\
person & cloth & & & &\\
\subfigure{\includegraphics[width=0.16\linewidth]{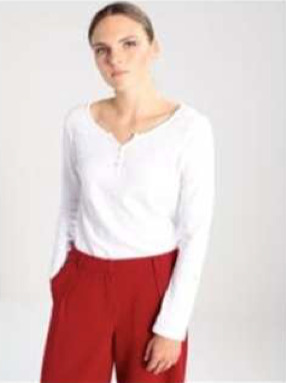}} &
\subfigure{\includegraphics[width=0.16\linewidth]{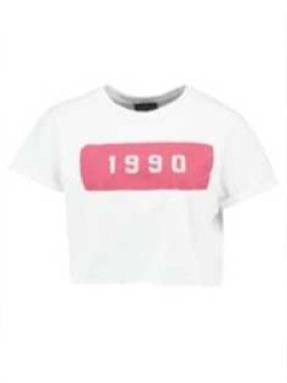}} &
\subfigure{\includegraphics[width=0.16\linewidth]{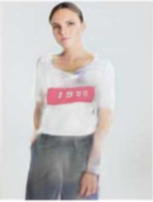}} &
\subfigure{\includegraphics[width=0.16\linewidth]{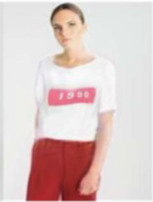}} &
\subfigure{\includegraphics[width=0.16\linewidth]{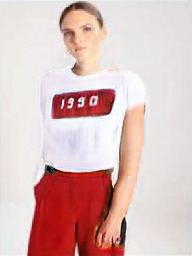}} &
\subfigure{\includegraphics[width=0.16\linewidth]{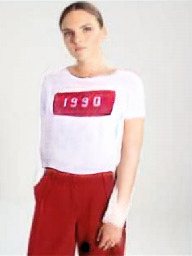}}\\ [-2.2ex]
\subfigure{\includegraphics[width=0.16\linewidth]{VTNFP/person_7.jpg}} &
\subfigure{\includegraphics[width=0.16\linewidth]{VTNFP/cloth_7.jpg}} &
\subfigure{\includegraphics[width=0.16\linewidth]{VTNFP/person_7_cloth_7_CPVTON.jpg}} &
\subfigure{\includegraphics[width=0.16\linewidth]{VTNFP/person_7_cloth_7_VT.jpg}} &
\subfigure{\includegraphics[width=0.16\linewidth]{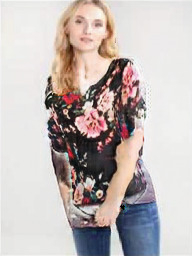}} &
\subfigure{\includegraphics[width=0.16\linewidth]{VTNFP/cloth_7person_7.jpg}}\\ [-2.2ex]
\subfigure{\includegraphics[width=0.16\linewidth]{VTNFP/person_8.jpg}} &
\subfigure{\includegraphics[width=0.16\linewidth]{VTNFP/cloth_8.jpg}} &
\subfigure{\includegraphics[width=0.16\linewidth]{VTNFP/person_8_cloth_8_CP_VTON.jpg}} &
\subfigure{\includegraphics[width=0.16\linewidth]{VTNFP/person_8_cloth_8_VT.jpg}} &
\subfigure{\includegraphics[width=0.16\linewidth]{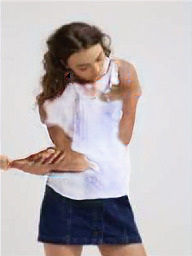}} &
\subfigure{\includegraphics[width=0.16\linewidth]{VTNFP/cloth_8person_8.jpg}}\\
\end{tabular*}

\end{figure}
\addtolength{\tabcolsep}{6pt}

\addtolength{\tabcolsep}{-6pt}
\begin{figure}[]
\ContinuedFloat
\fontsize{8}{8}\selectfont
\centering
\begin{tabular*}{340pt}{@{\extracolsep{\fill}}cccccc}
Reference & Target & CP-VTON & VTNFP & T-WUTON & S-WUTON\\
person & cloth & & & &\\
\subfigure{\includegraphics[width=0.16\linewidth]{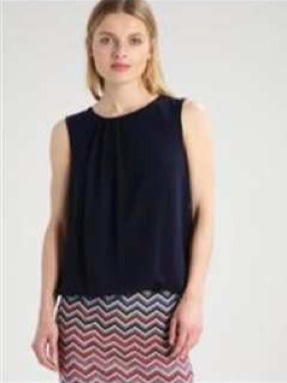}} &
\subfigure{\includegraphics[width=0.16\linewidth]{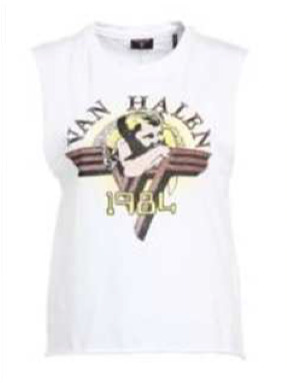}} &
\subfigure{\includegraphics[width=0.16\linewidth]{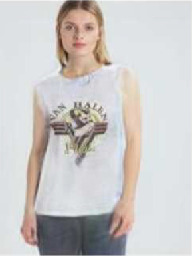}} &
\subfigure{\includegraphics[width=0.16\linewidth]{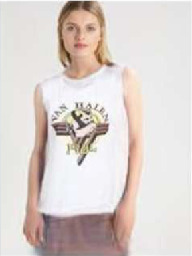}} &
\subfigure{\includegraphics[width=0.16\linewidth]{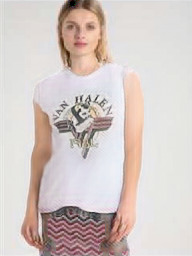}} &
\subfigure{\includegraphics[width=0.16\linewidth]{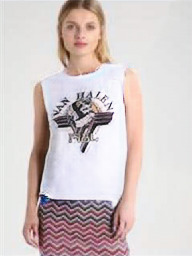}} \\ [-2.2ex]
\subfigure{\includegraphics[width=0.16\linewidth]{VTNFP/person_2.jpg}} &
\subfigure{\includegraphics[width=0.16\linewidth]{VTNFP/cloth_2.jpg}} &
\subfigure{\includegraphics[width=0.16\linewidth]{VTNFP/person_2_cloth_2_CPVTON.jpg}} &
\subfigure{\includegraphics[width=0.16\linewidth]{VTNFP/person_2_cloth_2_VT.jpg}} &
\subfigure{\includegraphics[width=0.16\linewidth]{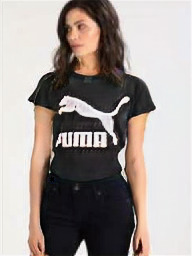}} &
\subfigure{\includegraphics[width=0.16\linewidth]{VTNFP/cloth_2person_2.jpg}} \\ [-2.2ex]
\subfigure{\includegraphics[width=0.16\linewidth]{VTNFP/person_3.jpg}} &
\subfigure{\includegraphics[width=0.16\linewidth]{VTNFP/cloth_3.jpg}} &
\subfigure{\includegraphics[width=0.16\linewidth]{VTNFP/person_3_cloth_3_CPVTON.jpg}} &
\subfigure{\includegraphics[width=0.16\linewidth]{VTNFP/person_3_cloth_3_VT.jpg}} &
\subfigure{\includegraphics[width=0.16\linewidth]{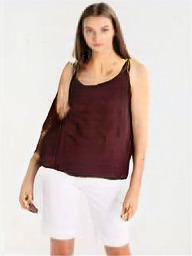}} &
\subfigure{\includegraphics[width=0.16\linewidth]{VTNFP/cloth_3person_3.jpg}} \\ [-2.2ex]
\subfigure{\includegraphics[width=0.16\linewidth]{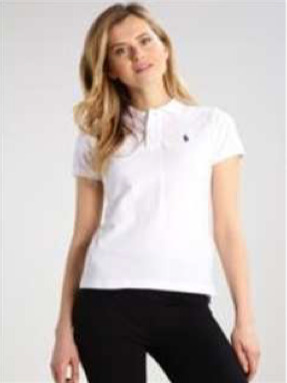}} &
\subfigure{\includegraphics[width=0.16\linewidth]{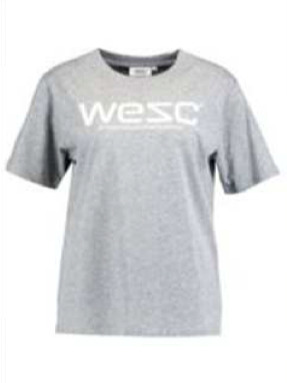}} &
\subfigure{\includegraphics[width=0.16\linewidth]{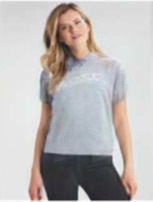}} &
\subfigure{\includegraphics[width=0.16\linewidth]{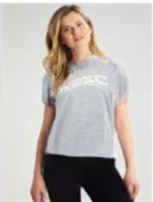}} &
\subfigure{\includegraphics[width=0.16\linewidth]{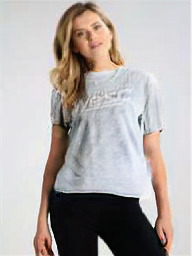}} &
\subfigure{\includegraphics[width=0.16\linewidth]{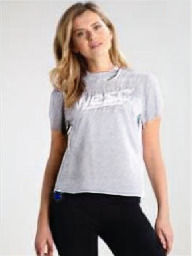}} \\ [-2.2ex]
\subfigure{\includegraphics[width=0.16\linewidth]{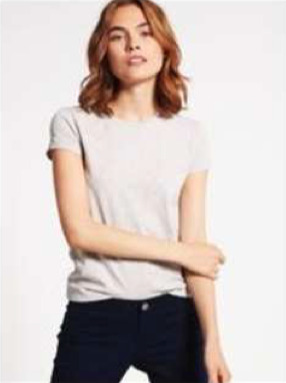}} &
\subfigure{\includegraphics[width=0.16\linewidth]{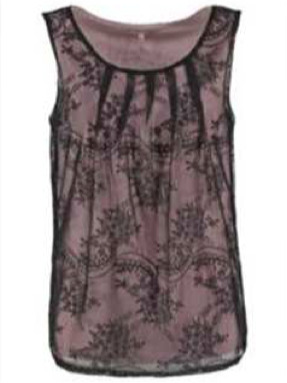}} &
\subfigure{\includegraphics[width=0.16\linewidth]{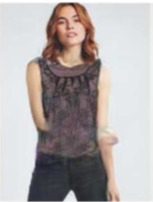}} &
\subfigure{\includegraphics[width=0.16\linewidth]{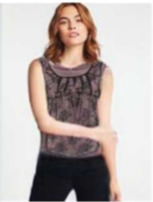}} &
\subfigure{\includegraphics[width=0.16\linewidth]{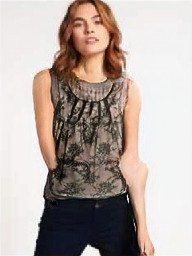}} &
\subfigure{\includegraphics[width=0.16\linewidth]{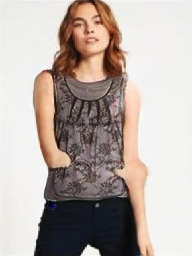}} \\ [-2.2ex]
\end{tabular*}
\caption{\label{fig:appendix_VTNFP} Comparisons of S-WUTON and VTNFP. Images are taken from VTNFP's paper, except for T-WUTON's and S-WUTON's columns. }
\end{figure}
\addtolength{\tabcolsep}{6pt}

\newpage
\subsection{Comparisons with images from ClothFlow \cite{Han_2019_ICCV} paper}
\begin{figure*}[]
\fontsize{8}{8}\selectfont
\centering
\begin{tabular*}{340pt}{@{\extracolsep{\fill}}ccccccc}
Reference & Target & CP-VTON & ClothFlow & T-WUTON & S-WUTON\\
person & cloth & & & & \\
\subfigure{\includegraphics[width=0.16\linewidth]{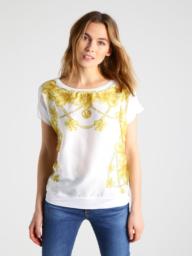}} &
\subfigure{\includegraphics[width=0.16\linewidth]{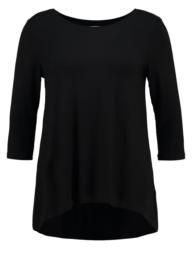}} &
\subfigure{\includegraphics[width=0.16\linewidth]{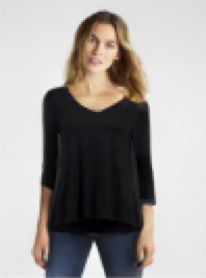}} &
\subfigure{\includegraphics[width=0.16\linewidth]{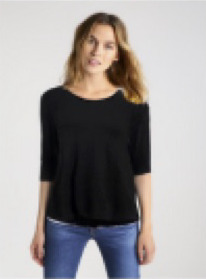}} &
\subfigure{\includegraphics[width=0.16\linewidth]{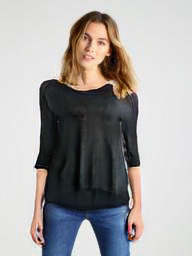}} &
\subfigure{\includegraphics[width=0.16\linewidth]{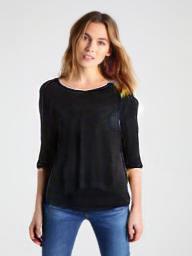}} \\ [-2.2ex]
\subfigure{\includegraphics[width=0.16\linewidth]{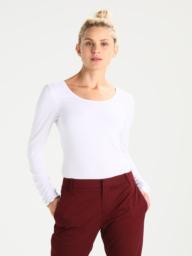}} &
\subfigure{\includegraphics[width=0.16\linewidth]{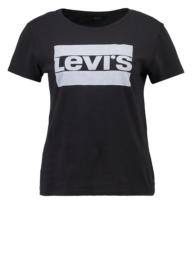}} &
\subfigure{\includegraphics[width=0.16\linewidth]{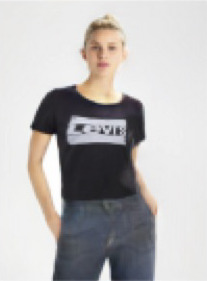}} &
\subfigure{\includegraphics[width=0.16\linewidth]{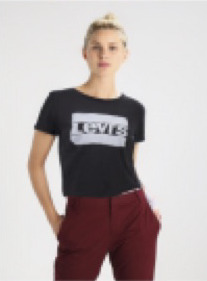}} &
\subfigure{\includegraphics[width=0.16\linewidth]{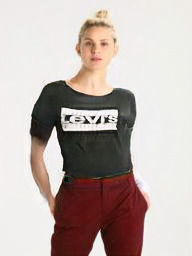}} &
\subfigure{\includegraphics[width=0.16\linewidth]{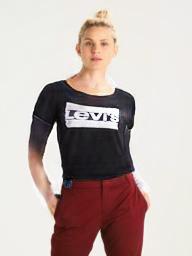}} \\ [-2.2ex]
\subfigure{\includegraphics[width=0.16\linewidth]{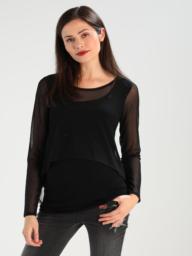}} &
\subfigure{\includegraphics[width=0.16\linewidth]{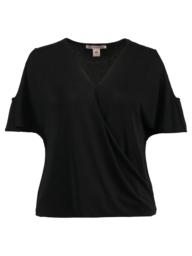}} &
\subfigure{\includegraphics[width=0.16\linewidth]{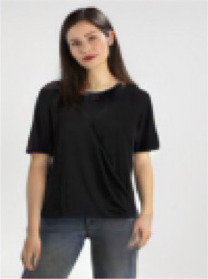}} &
\subfigure{\includegraphics[width=0.16\linewidth]{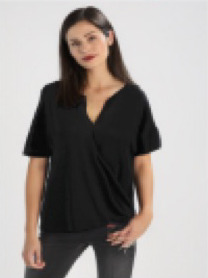}} &
\subfigure{\includegraphics[width=0.16\linewidth]{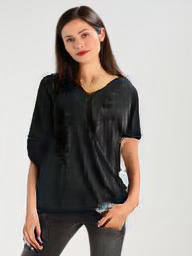}} &
\subfigure{\includegraphics[width=0.16\linewidth]{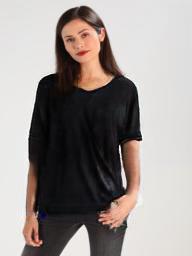}} \\ [-2.2ex]
\subfigure{\includegraphics[width=0.16\linewidth]{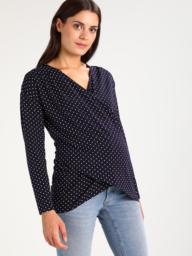}} &
\subfigure{\includegraphics[width=0.16\linewidth]{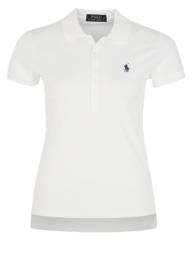}} &
\subfigure{\includegraphics[width=0.16\linewidth]{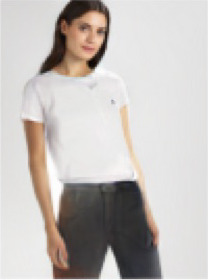}} &
\subfigure{\includegraphics[width=0.16\linewidth]{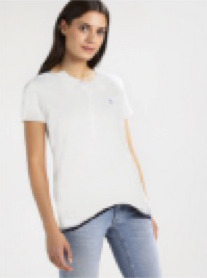}} &
\subfigure{\includegraphics[width=0.16\linewidth]{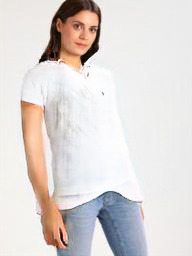}} &
\subfigure{\includegraphics[width=0.16\linewidth]{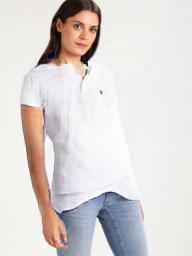}} \\ [-2.2ex]
\subfigure{\includegraphics[width=0.16\linewidth]{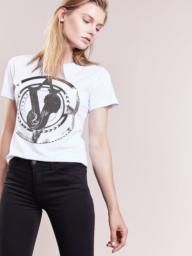}} &
\subfigure{\includegraphics[width=0.16\linewidth]{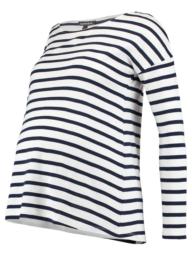}} &
\subfigure{\includegraphics[width=0.16\linewidth]{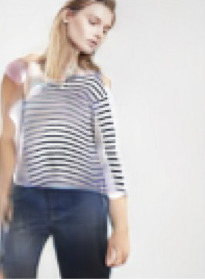}} &
\subfigure{\includegraphics[width=0.16\linewidth]{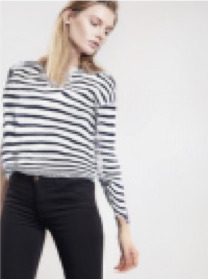}} &
\subfigure{\includegraphics[width=0.16\linewidth]{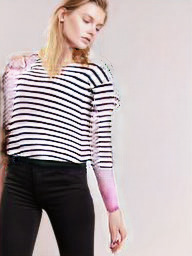}} &
\subfigure{\includegraphics[width=0.16\linewidth]{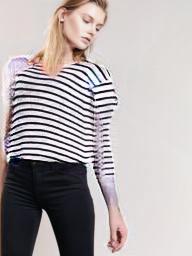}} \\ [-2.2ex]
\subfigure{\includegraphics[width=0.16\linewidth]{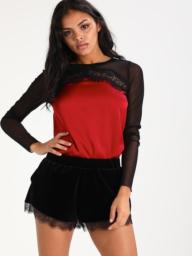}} &
\subfigure{\includegraphics[width=0.16\linewidth]{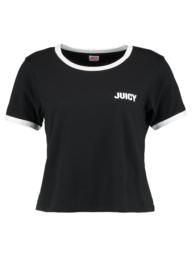}} &
\subfigure{\includegraphics[width=0.16\linewidth]{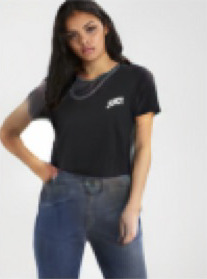}} &
\subfigure{\includegraphics[width=0.16\linewidth]{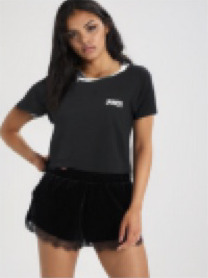}} &
\subfigure{\includegraphics[width=0.16\linewidth]{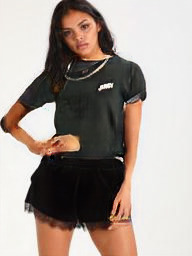}} &
\subfigure{\includegraphics[width=0.16\linewidth]{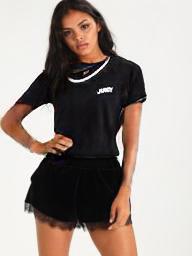}} \\ [-2.2ex]
\end{tabular*}
\caption{ \label{fig:appendix_clothflow} Comparisons of S-WUTON and ClothFlow. Images are taken from ClothFlow's paper, except for T-WUTON's and S-WUTON's columns. }
\end{figure*}

In Fig. \ref{fig:appendix_clothflow}, we show visual comparisons between CP-VTON, ClothFlow, T-WUTON and S-WUTON. Images from all columns except T-WUTON and S-WUTON are taken from their paper. Note that they trained their model on the original VITON dataset, that is now forbidden due to copyright issues. As mentioned in the paper, our model is trained on the dataset released in MG-VTON \cite{dong2019towards}.

We can observe that S-WUTON preserves better the shape of the hands of the person (rows 6,7,12).

Compared to ClothFlow, S-WUTON handles as well complex geometric deformations. S-WUTON seems to be slightly better on stripes (rows 4 and 11). Indeed, ClothFlow uses a dense flow to warp clothes, which means the warping module warps from the source to the target pixel-by-pixel. It has thus difficulties to keep the stripes straight.

In terms of computational cost at inference time, our S-WUTON is at least 13 times faster than their model. Moreover, since their pipeline relies on human parsing and pose estimation, it is also sensitive to the errors exhibited in our paper and in Fig. \ref{fig:appendix_clothflow} of the Appendix.

\addtolength{\tabcolsep}{-6pt}
\begin{figure}[]
\fontsize{8}{8}\selectfont
\centering
\begin{tabular*}{340pt}{@{\extracolsep{\fill}}cccccc}
Reference & Target & CP-VTON & ClothFlow & T-WUTON & S-WUTON\\
person & cloth & & & & \\
\subfigure{\includegraphics[width=0.16\linewidth]{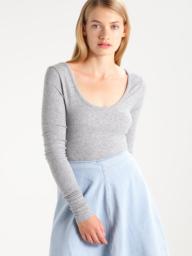}} &
\subfigure{\includegraphics[width=0.16\linewidth]{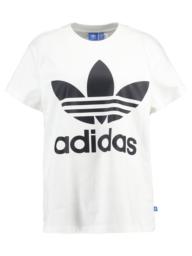}} &
\subfigure{\includegraphics[width=0.16\linewidth]{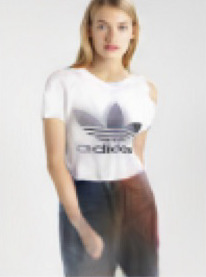}} &
\subfigure{\includegraphics[width=0.16\linewidth]{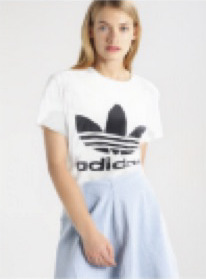}} &
\subfigure{\includegraphics[width=0.16\linewidth]{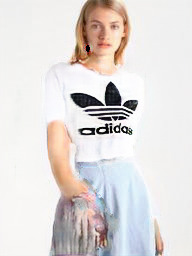}} &
\subfigure{\includegraphics[width=0.16\linewidth]{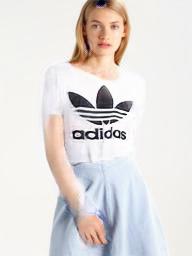}} \\ [-2.2ex]
\subfigure{\includegraphics[width=0.16\linewidth]{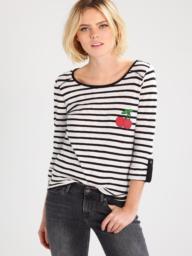}} &
\subfigure{\includegraphics[width=0.16\linewidth]{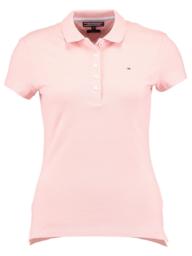}} &
\subfigure{\includegraphics[width=0.16\linewidth]{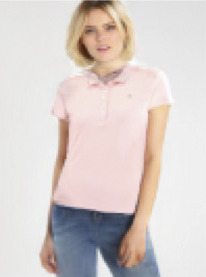}} &
\subfigure{\includegraphics[width=0.16\linewidth]{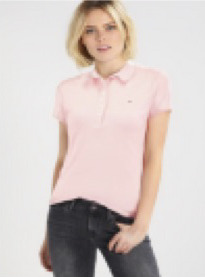}} &
\subfigure{\includegraphics[width=0.16\linewidth]{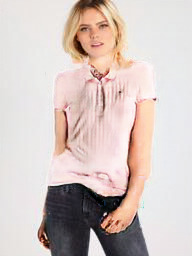}} &
\subfigure{\includegraphics[width=0.16\linewidth]{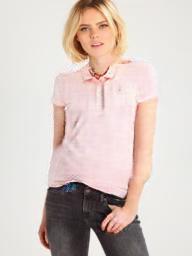}} \\ [-2.2ex]
\end{tabular*}
\end{figure}
\addtolength{\tabcolsep}{6pt}

\addtolength{\tabcolsep}{-6pt}
\begin{figure}[]
\fontsize{8}{8}\selectfont
\centering
\begin{tabular*}{340pt}{@{\extracolsep{\fill}}cccccc}
Reference & Target & CP-VTON & ClothFlow & T-WUTON & S-WUTON\\
person & cloth & & &  &\\
\subfigure{\includegraphics[width=0.16\linewidth]{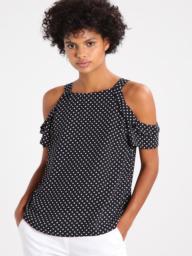}} &
\subfigure{\includegraphics[width=0.16\linewidth]{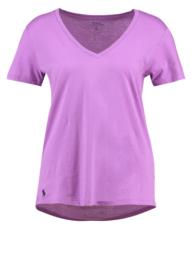}} &
\subfigure{\includegraphics[width=0.16\linewidth]{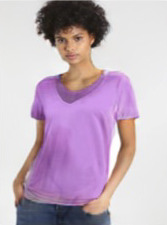}} &
\subfigure{\includegraphics[width=0.16\linewidth]{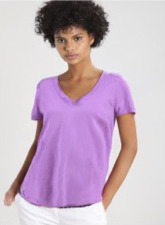}} &
\subfigure{\includegraphics[width=0.16\linewidth]{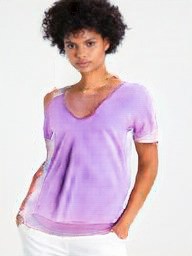}} &
\subfigure{\includegraphics[width=0.16\linewidth]{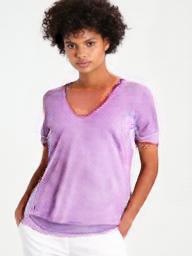}} \\ [-2.2ex]
\subfigure{\includegraphics[width=0.16\linewidth]{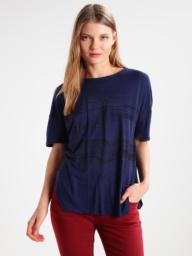}} &
\subfigure{\includegraphics[width=0.16\linewidth]{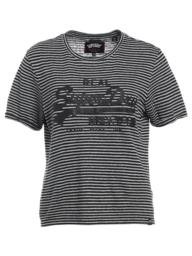}} &
\subfigure{\includegraphics[width=0.16\linewidth]{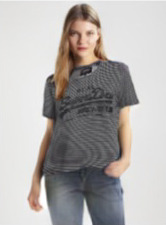}} &
\subfigure{\includegraphics[width=0.16\linewidth]{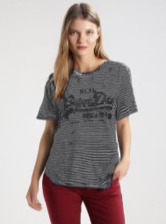}} &
\subfigure{\includegraphics[width=0.16\linewidth]{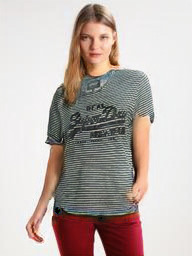}} &
\subfigure{\includegraphics[width=0.16\linewidth]{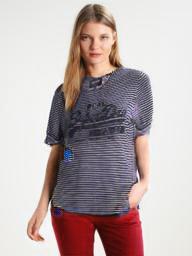}} \\ [-2.2ex]
\subfigure{\includegraphics[width=0.16\linewidth]{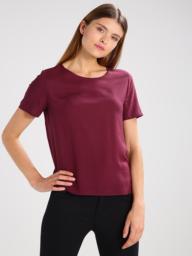}} &
\subfigure{\includegraphics[width=0.16\linewidth]{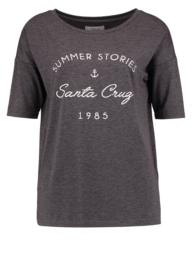}} &
\subfigure{\includegraphics[width=0.16\linewidth]{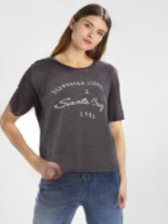}} &
\subfigure{\includegraphics[width=0.16\linewidth]{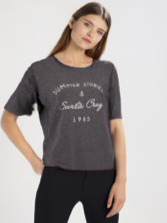}} &
\subfigure{\includegraphics[width=0.16\linewidth]{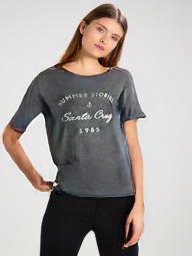}} &
\subfigure{\includegraphics[width=0.16\linewidth]{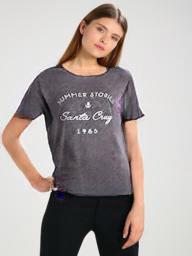}} \\ [-2.2ex]
\subfigure{\includegraphics[width=0.16\linewidth]{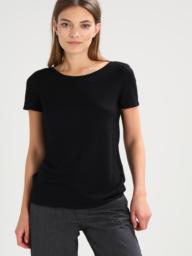}} &
\subfigure{\includegraphics[width=0.16\linewidth]{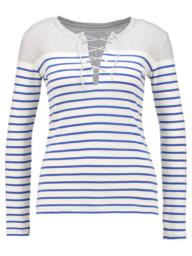}} &
\subfigure{\includegraphics[width=0.16\linewidth]{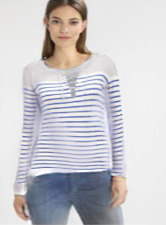}} &
\subfigure{\includegraphics[width=0.16\linewidth]{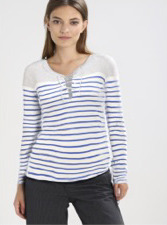}} &
\subfigure{\includegraphics[width=0.16\linewidth]{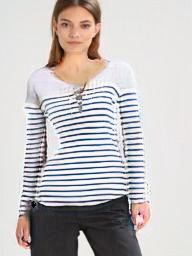}} &
\subfigure{\includegraphics[width=0.16\linewidth]{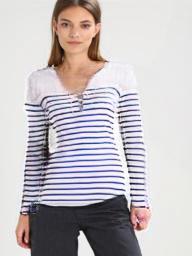}} \\ [-2.2ex]
\subfigure{\includegraphics[width=0.16\linewidth]{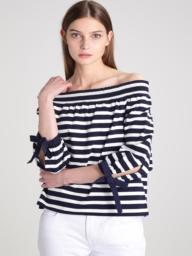}} &
\subfigure{\includegraphics[width=0.16\linewidth]{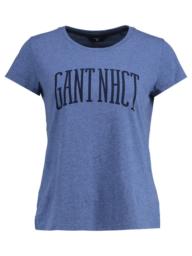}} &
\subfigure{\includegraphics[width=0.16\linewidth]{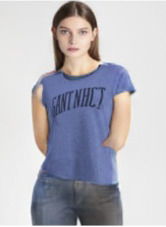}} &
\subfigure{\includegraphics[width=0.16\linewidth]{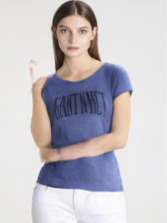}} &
\subfigure{\includegraphics[width=0.16\linewidth]{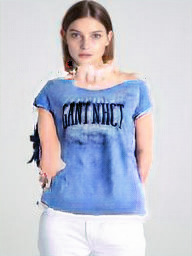}} &
\subfigure{\includegraphics[width=0.16\linewidth]{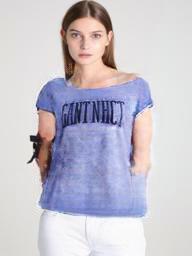}} \\ [-2.2ex]
\subfigure{\includegraphics[width=0.16\linewidth]{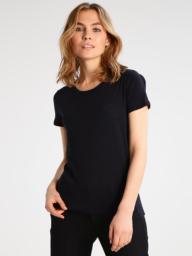}} &
\subfigure{\includegraphics[width=0.16\linewidth]{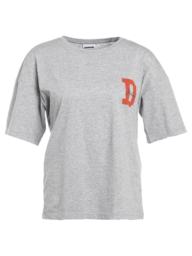}} &
\subfigure{\includegraphics[width=0.16\linewidth]{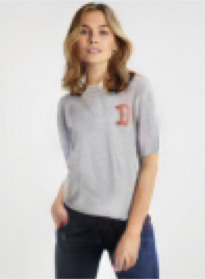}} &
\subfigure{\includegraphics[width=0.16\linewidth]{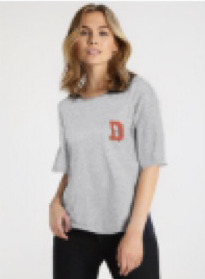}} &
\subfigure{\includegraphics[width=0.16\linewidth]{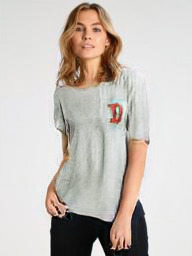}} &
\subfigure{\includegraphics[width=0.16\linewidth]{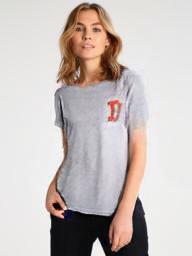}} \\ [-2.2ex]
\end{tabular*}
\end{figure}
\addtolength{\tabcolsep}{6pt}

\addtolength{\tabcolsep}{-6pt}
\addtolength{\tabcolsep}{6pt}

\newpage
\subsection{Ablation studies on T-WUTON}

\begin{figure*}[ht]
\fontsize{7}{7}\selectfont
\centering\begin{tabular}{c@{}c@{}c@{}c@{}c@{}c@{}c@{}c@{}c@{}}
Reference & Target & CP-VTON & T-WUTON & T-WUTON w. & T-WUTON & T-WUTON not\\
 person & cloth & & (ours)  & paired $L_{adv}$ & w/o $L_{adv}$ & end-to-end \\
\subfigure{\includegraphics[width=0.12\linewidth]{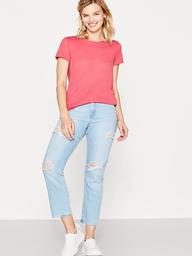}} &
\subfigure{\includegraphics[width=0.12\linewidth]{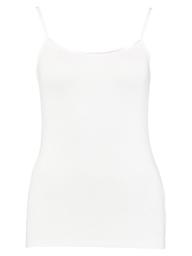}} &
\subfigure{\includegraphics[width=0.12\linewidth]{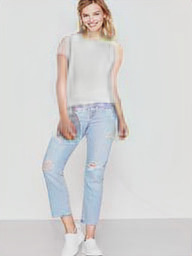}}&
\subfigure{\includegraphics[width=0.12\linewidth]{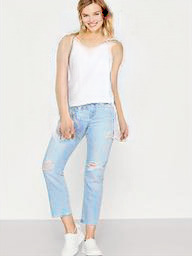}} &
\subfigure{\includegraphics[width=0.12\linewidth]{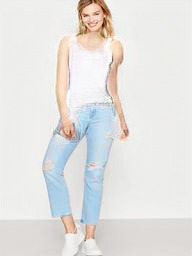}} &
\subfigure{\includegraphics[width=0.12\linewidth]{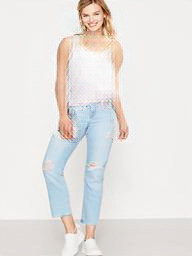}} &
\subfigure{\includegraphics[width=0.12\linewidth]{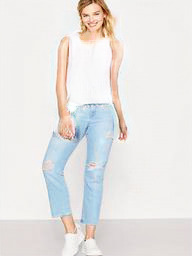}} \\[-2.3ex]
\subfigure{\includegraphics[width=0.12\linewidth]{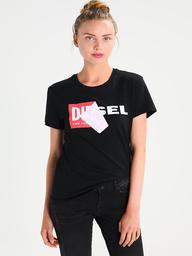}} &
\subfigure{\includegraphics[width=0.12\linewidth]{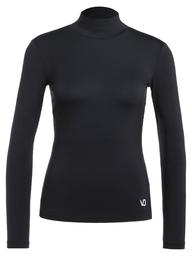}} &
\subfigure{\includegraphics[width=0.12\linewidth]{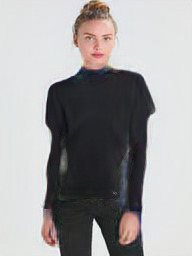}}&
\subfigure{\includegraphics[width=0.12\linewidth]{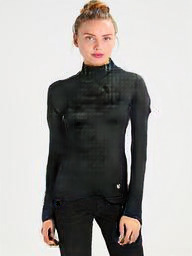}} &
\subfigure{\includegraphics[width=0.12\linewidth]{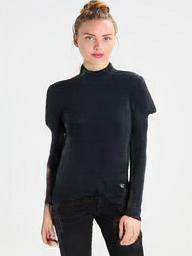}} &
\subfigure{\includegraphics[width=0.12\linewidth]{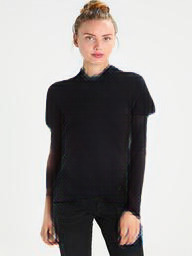}} &
\subfigure{\includegraphics[width=0.12\linewidth]{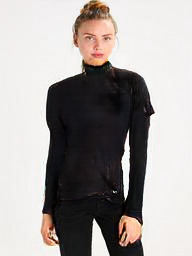}} \\[-2.3ex]
\subfigure{\includegraphics[width=0.12\linewidth]{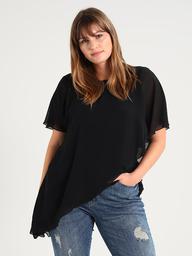}} &
\subfigure{\includegraphics[width=0.12\linewidth]{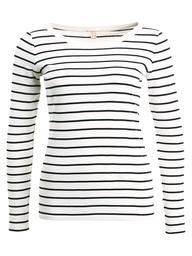}} &
\subfigure{\includegraphics[width=0.12\linewidth]{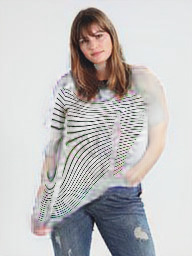}} &
\subfigure{\includegraphics[width=0.12\linewidth]{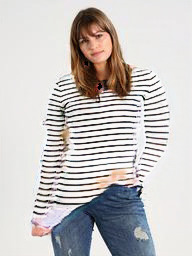}} &
\subfigure{\includegraphics[width=0.12\linewidth]{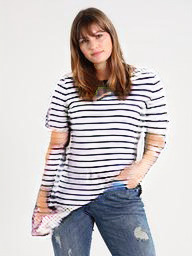}} &
\subfigure{\includegraphics[width=0.12\linewidth]{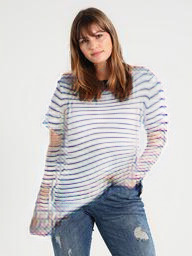}} &
\subfigure{\includegraphics[width=0.12\linewidth]{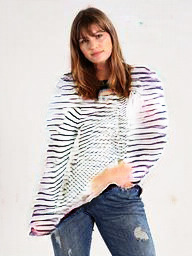}} \\[-2.3ex]
\subfigure{\includegraphics[width=0.12\linewidth]{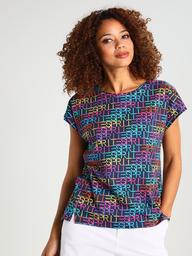}} &
\subfigure{\includegraphics[width=0.12\linewidth]{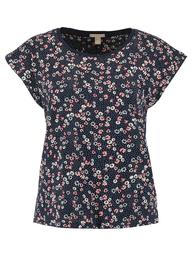}} &
\subfigure{\includegraphics[width=0.12\linewidth]{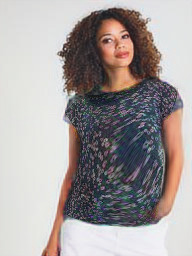}}&
\subfigure{\includegraphics[width=0.12\linewidth]{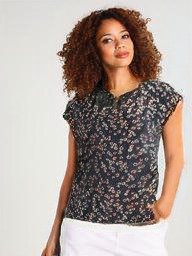}} &
\subfigure{\includegraphics[width=0.12\linewidth]{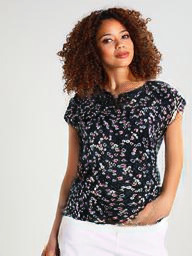}} &
\subfigure{\includegraphics[width=0.12\linewidth]{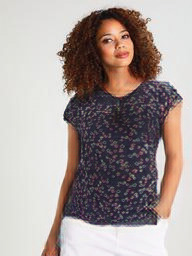}} &
\subfigure{\includegraphics[width=0.12\linewidth]{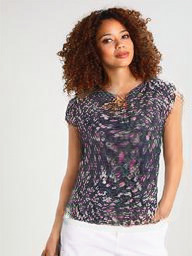}} \\
\end{tabular}
\caption{\label{fig:loss}Impact of loss functions on T-WUTON: The unpaired adversarial loss function improves the performance of T-WUTON in the case of significant shape changes from the source cloth to the target cloth. Specifically, when going from short sleeves to long sleeves, it tends to gum the shape of the short sleeves. With the paired adversarial loss, we do not observe this phenomenon since the case never happens during training. Without the adversarial loss, images are blurry and less sharp. Finally, the end-to-end training is key to realistic geometric deformations (see last column).}
\end{figure*}

\begin{table*}[h]
\begin{center}
\begin{tabular}{l|c|c|c|c}
Method & T-WUTON & W/o $L_{adv}$ & Paired $L_{adv}$ & Not end-to-end \\
\hline
Paired $L_{adv}$ &  & &  \checkmark &  \\ \hline
Unpaired $L_{adv}$ &  \checkmark & & & \checkmark  \\ \hline
End-to-end &  \checkmark &  \checkmark &  \checkmark &   \\ \hline
LPIPS & 0.101 $\pm$ 0.047 & 0.107 $\pm$ 0.049 & \textbf{0.099}  $\pm$ 0.046 & 0.112 $\pm$ 0.053 \\
SSIM & 0.799 $\pm$ 0.089 & 0.799 $\pm$ 0.088 & \textbf{0.800} $\pm$	0.089 &  0.799 $\pm$ 0.089 \\
IS & 3.114  $\pm$ 0.118 & 2.729 $\pm$	0.091 & 3.004 $\pm$	0.091 & 3.102 $\pm$	0.077 \\
FID & 9.877 & 13.020 & \textbf{8.298} & 11.125 \\
\end{tabular}
\end{center}
\caption{\label{table:lpips_ablation} Ablation studies on T-WUTON. Quantitative metrics on paired setting (LPIPS and SSIM) and on unpaired setting (IS and FID). For LPIPS and FID, the lower is the better. For SSIM and IS, the higher is the better. $\pm$ reports std. dev.}
\end{table*}
To investigate the effectiveness of T-WUTON's components, we perform several ablation studies.
In Fig. \ref{fig:loss}, we show visual comparisons of CP-VTON and different variants of our approach: T-WUTON; T-WUTON with an adversarial loss on paired data (\textit{i.e.} the adversarial loss is computed with the same synthesized image as the L1 and VGG losses); T-WUTON without the adversarial loss; T-WUTON without back-propagating the loss of the synthesized images ($L_1, L_{perceptual}, L_{adv}$) to the geometric matcher.

The results in Fig. \ref{fig:loss} as well as FID and LPIPS metrics in Table \ref{table:lpips_ablation} show the importance of our end-to-end learning of geometric deformations. When the geometric matcher only benefits from $L_{warp}$, it only learns to align $c$ with the masked area in $p^{\star}$. However, it does not preserve the inner structure of the cloth. Back-propagating the loss computed on the synthesized images $\Tilde{p}$ alleviates this issue. The quantitative results of IS and SSIM scores on the not end-to-end variant show that these metrics are less suited to the virtual try-on task than LPIPS.

The adversarial loss generates sharper images and improves the contrast. This is confirmed by the LPIPS, IS and FID metrics in Table \ref{table:lpips_ablation} and with visual results in Fig. \ref{fig:loss}. With the unpaired adversarial setting, the system better handles large variations between the shape of the cloth worn by the person and the shape of the new cloth. On metrics in the paired setting (LPIPS and SSIM), the best model is the variant using adversarial loss on paired data, which is logical. However, visual investigation suggests that the unpaired adversarial loss is better in the real use case of our work (see Fig. \ref{fig:loss}).

\end{document}